\documentclass[pdflatex,sn-mathphys-num,iicol]{sn-jnl}

\usepackage{graphicx}%
\usepackage{multirow}%
\usepackage{amsmath,amssymb,amsfonts}%
\usepackage{amsthm}%
\usepackage{mathrsfs}%
\usepackage[title]{appendix}%
\usepackage{xcolor}%
\usepackage{textcomp}%
\usepackage{manyfoot}%
\usepackage{booktabs}%
\usepackage{algorithm}%
\usepackage{algorithmicx}%
\usepackage{algpseudocode}%
\usepackage{listings}%
\usepackage{subcaption}

\newcommand{\rev}[1]{#1}
\newenvironment{revblock}{}{}
\colorlet{blue}{black}

\theoremstyle{thmstyleone}%
\theoremstyle{thmstyletwo}%

\theoremstyle{thmstylethree}%

\begin{document}

\title[selective prediction]{When should we trust the annotation? Selective prediction 
for molecular structure retrieval from mass spectra}

\author*[1]{\fnm{Mira} \sur{Jürgens}}\email{mira.juergens@ugent.be}

\author[2]{\fnm{Gaetan} \sur{De Waele}}\email{gaetan.dewaele@uantwerpen.be}
\author[1]{\fnm{Morteza} \sur{Rakhshaninejad}}\email{morteza.rakhshaninejad@ugent.be}
\author[1]{\fnm{Willem} \sur{Waegeman}}\email{willem.waegeman@ugent.be}

\affil*[1]{\orgdiv{Department of Data Analysis and Mathematical Modeling}, \orgname{Ghent University}, \orgaddress{\street{Coupure Links 653}, \city{Ghent}, \postcode{9000}, \country{Belgium}}}

\affil[2]{\orgdiv{Department of Computer Science}, \orgname{University of Antwerp}, \orgaddress{\street{Middelheimlaan 1}, \city{Antwerp}, \postcode{2020}, \country{Belgium}}}

\abstract{Machine learning methods for identifying molecular structures from tandem mass spectra (MS/MS) have advanced rapidly, 
yet current approaches still exhibit significant error rates. In high-stakes applications such as clinical metabolomics and environmental 
screening, incorrect annotations can have serious consequences, making it essential to determine when a prediction can be trusted.
 In this work, we introduce a selective prediction framework for molecular structure retrieval from MS/MS spectra,
  \rev{separating low-risk predictions automatically from lower-confidence predictions}.
   We formulate the problem within the risk-coverage tradeoff framework and \rev{systematically
   evaluate uncertainty quantification strategies at three levels}:
    \rev{input-level distance in the learned representation,} fingerprint-level uncertainty over predicted molecular fingerprint bits,
    and retrieval-level uncertainty over candidate rankings. We compare scoring functions including first-order confidence measures, aleatoric and
    epistemic uncertainty estimates from second-order distributions, as well as distance-based measures in the latent space.
     All experiments are conducted on the MassSpecGym benchmark.
      Our analysis reveals that while fingerprint-level uncertainty scores are poor proxies for retrieval success,
      retrieval-level total uncertainty provides the overall strongest rejection criterion, and first-order 
      confidence measures are computationally inexpensive, strong baselines.
      We demonstrate that by applying distribution-free risk control via generalisation bounds,
       practitioners can specify a tolerable error rate and obtain a subset of annotations satisfying that constraint with high probability.

\textbf{Scientific contribution}: \rev{This work provides a systematic evaluation of selective
prediction for molecular structure retrieval from mass spectra, quantifying uncertainty at
three levels of the retrieval pipeline: the predicted molecular fingerprint, the input embedding, and
the retrieval-level candidate ranking. Our central finding is that the most informative uncertainty is
the one whose level matches the target loss: retrieval-level scores are most effective when the target
is retrieval accuracy, whereas fingerprint-level uncertainty is most effective when the
target is fingerprint similarity. We show how
distribution-free risk control can be applied to obtain provable finite-sample guarantees on annotation quality,
turning molecular identification into an uncertainty-aware decision process.}

}


\maketitle

\section{Introduction}
\label{sec: introduction}
\begin{figure*}[t]
  \centering
  \includegraphics[width=\textwidth]{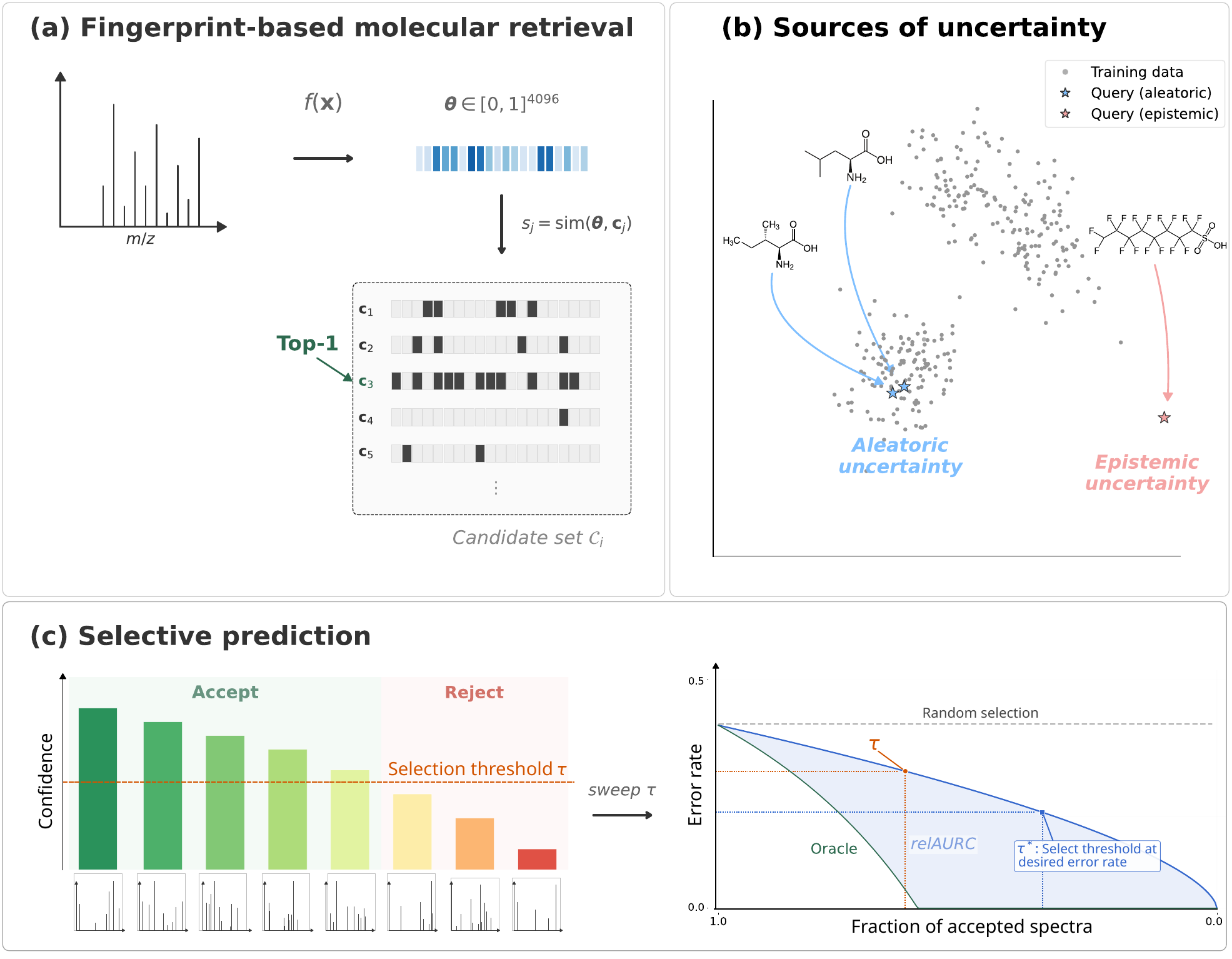}
  \caption{Overview of the selective prediction framework for molecular structure retrieval from tandem mass spectra.
  \textbf{(a)} Fingerprint-based molecular retrieval. A tandem mass spectrum $\boldsymbol{x}$ is mapped by a model $f$
  to a vector of bitwise probabilities $\boldsymbol{\theta}\in [0,1]^{4096}$. Each candidate fingerprint 
  $\boldsymbol{c}_j$ in the instance-specific candidate set $\mathcal{C}_i$ is scored by cosine similarity 
  $s_j = \mathrm{sim}(\boldsymbol{\theta}, \mathbf{c}_j)$, and candidates are ranked by descending score.
  \textbf{(b)} Sources of uncertainty in the learned representation space. Aleatoric uncertainty is due to 
  inherent noise and ambiguity in the data, and can arise for structurally similar molecules.
  Epistemic uncertainty reflects a general lack of knowledge, and arises in regions of the representation space that are far from training data.
  \textbf{(c)} Selective prediction. Left: test spectra are sorted by confidence in descending order.
  A threshold $\tau$ partitions spectra into accepted and rejected predictions. Right: sweeping the threshold $\tau$
  over its full range traces the risk-coverage curve, where different points on the curve correspond to different risk-coverage trade-offs.
  } 
  \label{fig: figure1}
\end{figure*}

Untargeted metabolomics generates vast numbers of tandem mass spectra (MS/MS), yet only approximately $10\%$ of detected features
can be annotated with molecular structures \cite{bocker2016fragmentation,hahnke2018pubchem}, a deficit also known as 
\textit{dark matter} of metabolomics \cite{da2015illuminating,bittremieux2022critical}.
This difficulty stems from different sources, including a fundamental mismatch between the vast chemical space and
 the limited coverage of spectral reference libraries,
the inherent complexity of molecular fragmentation, and the variability of spectra across instrument settings and ionization conditions.
The correct identification
of the molecular structure from a mass spectrum is hence considered a very challenging task, 
while being of fundamental importance with applications in drug discovery \cite{aebersold2003mass,meissner2022emerging,skinnider2021deep},
environmental biochemistry \cite{pico2020pyrolysis,hollender2017nontarget} and clinical diagnostics \cite{miller2015untargeted}.
A growing body of literature uses machine learning methods to tackle this problem, where two main paradigms are formed by
\textit{molecular retrieval} \cite{duhrkop2015searching,wei2019rapid,goldman2023annotating,kalia2025jestr} and 
\textit{de novo structure elucidation} \cite{stravs2022msnovelist,butler2023ms2mol,bohdediffms,hanms}.
The retrieval paradigm, which is the focus of this work, predicts a molecular representation
 from the query spectrum and ranks candidate structures from a molecular database via a similarity function.
The classical approach within this paradigm is molecular fingerprint prediction \cite{duhrkop2015searching,wei2019rapid,goldman2023annotating},
where a model learns to map spectra to binary fingerprint vectors encoding the presence or absence of molecular substructures \cite{rogers2010extended,bajusz2015tanimoto}.
More recently, joint embedding approaches avoid explicit fingerprint prediction
and instead learn a shared latent space in which spectra and molecules can be compared directly via cosine similarity \cite{kalia2025jestr,xie2025csu}.

The pace of progress in retrieval-based identification has been substantial.
CSI:FingerID~\cite{duhrkop2015searching} trains an array of support vector machines, one per molecular property, using
multiple kernel learning over fragmentation tree similarities to predict molecular fingerprints from MS/MS spectra.
 It was the first method to substantially outperform in silico fragmentation tools such as MetFrag \cite{ruttkies2016metfrag}
and CFM-ID \cite{allen2014cfm}, and became the core of the widely adopted SIRIUS framework~\cite{duhrkop2019sirius}.
\rev{Building on this pipeline, COSMIC~\cite{hoffmann2022high} introduced supervised confidence scoring
for the top-ranked CSI:FingerID annotation of each spectrum, combining calibrated CSI:FingerID scores
with candidate-list, fragmentation-tree and fingerprint features to prioritize high-confidence hits.}
MIST~\cite{goldman2023annotating} subsequently introduced a transformer architecture that encodes
MS/MS peaks as chemical formula vectors rather than binned $m/z$ values, thereby implicitly featurizing
pairwise neutral losses between fragments. The introduction of the MassSpecGym benchmark~\cite{bushuiev2024massspecgym},
comprising 231{,}104 labeled spectra with structure-disjoint evaluation splits based on molecular similarity,
provided a standardized basis for comparing retrieval methods. Joint embedding approaches, which embed molecules
 and spectra in a shared latent space and rank candidates by cosine similarity
rather than by comparing explicit fingerprint reconstructions, have since demonstrated substantial improvements.
JESTR~\cite{kalia2025jestr} treats molecules and their spectra as two views of the same underlying object 
and trains molecular and spectral encoders jointly using contrastive multiview coding \cite{tian2020contrastive}.
GMLR \cite{zhang2025breaking} initializes the molecular encoder from ChemFormer \cite{irwin2022chemformer}
and jointly finetunes it during contrastive training.

Despite this rapid progress, a significant error rate at the current state of the art remains consequential for downstream applications.
In clinical metabolomics, incorrect identifications can delay or misdirect diagnosis of inborn errors of the metabolism~\cite{miller2015untargeted}.
In environmental nontarget screening, annotations directly inform regulatory decisions on chemical contamination~\cite{hollender2017nontarget}.
These examples illustrate that predictive accuracy alone, without a mechanism to express the reliability of individual predictions,
is insufficient for trustworthy metabolomics.
To avoid making unreliable predictions, correct uncertainty quantification can help to build 
a trustworthy deployment of machine learning pipelines. \textit{Selective prediction} \cite{pudil1992multistage,chow2009optimum,el2010foundations}
provides a powerful tool to reduce the error rate by abstaining from making a prediction when the uncertainty
is too high. A selective classifier augments a prediction function with a selection function that determines,
for each input, whether to predict or to abstain, thereby trading off \textit{coverage}, i.e.\ the fraction of 
inputs on which the model commits, for lower \textit{risk} on accepted predictions~\cite{el2010foundations,geifman2017selective}.

The effectiveness of selective prediction depends on the scoring function
that decides which predictions to trust.
A natural source of such scores is predictive uncertainty, which can be
decomposed into \textit{aleatoric} uncertainty, arising from irreducible
noise or ambiguity in the data, and \textit{epistemic} uncertainty,
reflecting the model's lack of knowledge due to limited
training data \cite{hullermeier2021aleatoric,kendall2017uncertainties}.
In mass spectrometry, aleatoric uncertainty is inherent to the
measurement process: structural isomers can produce nearly
indistinguishable fragmentation
patterns \cite{xu2015avoiding}, and spectra of the
same compound vary with collision energy, instrument type, and
acquisition conditions.
Epistemic uncertainty on the other hand reflects limitations of the learned model,
arising from finite training data, model misspecification,
and distributional shift between train and test data.
Until now, no empirical consensus exists for which of these uncertainty types is most informative for selective prediction.
We therefore analyse both components separately throughout this work.

In this work, we introduce a systematic selective prediction framework for molecular structure retrieval from MS/MS spectra (Figure~\ref{fig: figure1}).
We formulate the retrieval problem within the risk-coverage tradeoff framework and \rev{evaluate
uncertainty quantification strategies at three levels:
\textit{input-level} distance in the learned representation, \textit{bitwise} uncertainty over predicted
molecular fingerprint bits, and \textit{retrieval-level} uncertainty over candidate rankings. We compare
first-order confidence measures, epistemic uncertainty estimates from posterior approximations, and
distance-based measures in the learned representation space.}
To provide finite-sample guarantees, we apply distribution-free risk control
via the SGR algorithm \cite{geifman2017selective},
\rev{yielding provable bounds on the risk among accepted annotations under exchangeability.}
Our analysis reveals that while standard epistemic uncertainty estimates fail to improve selective prediction performance
in this setting,
computationally inexpensive first-order confidence measures based on retrieval
provide effective selection functions.
We demonstrate that selective prediction enables practitioners to specify a tolerable error rate
and obtain a subset of annotations satisfying that constraint with high probability,
transforming molecular identification into an uncertainty-aware decision process.
Section~\ref{sec:methods} introduces the main framework, uncertainty estimation and evaluation methods,
Section~\ref{sec: results and discussion} presents the experimental evaluation and discussion,
and Section~\ref{sec: conclusion} concludes with implications and limitations.

\section{Methods}

\label{sec:methods}
\subsection{Problem setting: fingerprint-based molecular retrieval}
\label{sec:problem_setting}
Let $\mathcal{X}$ denote the input space of tandem mass spectra and let
$\mathcal{Y} \subseteq \{0,1\}^{D}$ denote the output space of molecular fingerprints,
where $D\in \mathbb{N}$ is the fingerprint dimensionality.
Each fingerprint is a binary vector encoding the presence or absence of
predefined molecular substructures.
We assume access to a dataset
$\mathcal{D} = \{(\mathbf{x}_i, \mathbf{y}_i, \mathcal{C}_i)\}_{i=1}^{N}$
of $N$ labeled instances, where $\mathbf{x}_i \in \mathcal{X}$ is a tandem mass
spectrum, $\mathbf{y}_i \in \mathcal{Y}$ the molecular fingerprint of the true molecule, and
$\mathcal{C}_i = \{\mathbf{c}_{i,1}, \ldots, \mathbf{c}_{i,M_i}\} \subset \mathcal{Y}$
an instance-specific candidate set of $M_i$ molecular fingerprints retrieved from
a chemical database.
By construction, the true fingerprint satisfies
$\mathbf{y}_i \in \mathcal{C}_i$.

Molecular identification proceeds in two stages.
First, a model $f : \mathcal{X} \to \Theta \subseteq [0,1]^{D}$ predicts a vector of bitwise
probabilities $\boldsymbol{\theta} = (\theta_1, \dots, \theta_D)^T$, where each component
$\theta_d$ estimates the probability of the $d$-th substructure being
present.
Second, a similarity function $\mathrm{sim} : [0,1]^{D} \times \{0,1\}^{D} \to \mathbb{R}$
is used to score each candidate $\mathbf{c} \in \mathcal{C}_i$ against the
prediction.
In this work we use the cosine similarity,
\begin{equation}\label{eq:cosine_sim}
  \begin{aligned}
    s_j(\mathbf{x})
    &= \mathrm{sim}(\boldsymbol{\theta}, \mathbf{c}_{j}) \\
    &= \frac{\boldsymbol{\theta}^{\top} \mathbf{c}_{j}}
            {\|\boldsymbol{\theta}\|_2 \, \|\mathbf{c}_{j}\|_2}\,,
       \qquad j = 1, \ldots, M_i\,.
  \end{aligned}
\end{equation}
Candidates are ranked by descending score, and the top-$K$ set
$\mathcal{S}_K(\mathbf{x}) \subset \mathcal{C}_i$ consists of the $K$ highest-scoring
candidates.
Retrieval performance is measured by the \emph{hit rate at $K$},
\begin{equation}
    \label{eq:hitrate}
  \mathrm{Hit@}K(\mathbf{x}, \mathbf{y})
  = \mathbb{I}\!\left[\mathbf{y} \in \mathcal{S}_K(\mathbf{x})\right],
\end{equation}
where $\mathbb{I}[\cdot]$ denotes the indicator function, which equals one if the condition is satisfied and zero otherwise.
This metric captures the practical utility of the retrieval pipeline, as it directly reflects the probability
of the true molecule being among the top-$K$ candidates.
We evaluate $K \in \{1, 5, 20\}$ throughout this work.

Two properties of this setting are particularly relevant for uncertainty
quantification.
First, the candidate set size $M_i$ varies across instances, since it is
determined by external database queries based on precursor mass or molecular
formula.
Second, the task exhibits a two-stage structure in which 
retrieval quality directly depends on the quality of the predicted fingerprint. Whether uncertainty
at the bit level transfers to the retrieval level is analysed in Section \ref{sec: results and discussion}.

\subsection{Selective prediction}
\label{sec:selective_prediction}

In many applications, making no prediction is preferable to making an
incorrect one \cite{chow2009optimum,el2010foundations}.
Selective prediction formalizes this idea by augmenting a predictor $f$ with a
\emph{selection function} $g : \mathcal{X} \to \{0, 1\}$ that decides, for each
input, whether to commit to the prediction or to
abstain.
The resulting selective classifier $(f, g)$ operates as
\begin{equation}\label{eq:selective_classifier}
  (f, g)(\mathbf{x}) =
  \begin{cases}
    f(\mathbf{x}), & \text{if } g(\mathbf{x}) = 1, \\
    \text{abstain}, & \text{if } g(\mathbf{x}) = 0.
  \end{cases}
\end{equation}
In practice, the selection function is defined via a \emph{scoring function}
$\kappa : \mathcal{X} \to \mathbb{R}$ that quantifies the model's confidence in
its prediction, together with a threshold $\tau \in \mathbb{R}$:
\begin{equation}
\label{eq:selection_function}
  g_\tau(\mathbf{x}) = \mathbb{I}\!\left[\kappa(\mathbf{x}) \geq \tau\right].
\end{equation}
Instances for which the confidence falls below $\tau$ are rejected.
\rev{Here, rejecting an instance means that it is not accepted for automated annotation. The
prediction is not deleted, but can be kept as a lower-confidence candidate that can be handled
separately, for instance by an expert.}
Different choices of $\kappa$ are compared in Section~\ref{sec:scoring-functions}.
The quality of a selective classifier is characterized by two quantities.
The \emph{coverage} measures the fraction of instances on which the model
commits to a prediction,
\begin{equation}
\label{eq:coverage}
  \varphi(f, g) = \mathbb{E}[g(\mathbf{x})],
\end{equation}
and the \emph{selective risk} measures the expected loss restricted to accepted
instances,
\begin{equation}\label{eq:selective_risk}
  R(f, g) = \frac{\mathbb{E} [\ell(f(\mathbf{x}), \mathbf{y}) \cdot
  g(\mathbf{x})]}{\varphi(f, g)},
\end{equation}
where $\ell: \Theta \times \mathcal{Y} \rightarrow \mathbb{R}$ denotes a task-specific loss function.
For the retrieval task, we define
$\ell_K(f(\mathbf{x}), \mathbf{y}) = 1 - \mathrm{Hit@}K(\mathbf{x}, \mathbf{y})$,
so that $R(f, g)$ represents the miss rate among accepted predictions.
The central trade-off in selective prediction is between risk and coverage:
a more stringent threshold $\tau$ reduces the risk among selected instances
but also lowers coverage. This trade-off is visualized through
\emph{risk-coverage curves}, which show the selective risk as a function of coverage
for varying $\tau$.

\subsection{Scoring functions}
\label{sec:scoring-functions}
The effectiveness of selective prediction depends critically on the
scoring function~$\kappa: \mathcal{X} \rightarrow \mathbb{R}$ that drives the selection function in
Eqn.~\eqref{eq:selection_function}.
A well-chosen $\kappa$ assigns high values to inputs where the expected
risk is low, enabling a favorable risk-coverage trade-off.
Since the fingerprint-based retrieval pipeline operates in two stages,
uncertainty can arise at different levels, and it is a priori
unclear which characterization yields the most informative selection
criterion.
We therefore investigate scoring functions from three complementary
families \rev{(summarised in Table~\ref{tab:scoring-functions})}, distinguished by the level on which uncertainty is measured:
\textit{fingerprint-level} scores derived from the predicted bit
probabilities,
\textit{retrieval-level} scores operating on the induced candidate
ranking, and \textit{distance-based} scores that measure proximity
to the training distribution in representation space.
Within the first two families, we further distinguish between
\textit{first-order} scores that can be computed from a single
model prediction, and \textit{second-order} scores that require
an approximate posterior distribution over model parameters.
All $\kappa$ are oriented so that higher values indicate greater
confidence.
Formal definitions of all scoring functions are given in
Appendix~\ref{appendix: uncertainty metrics}. 

\bmhead{Candidate probabilities}
Several scoring functions operate not on the raw similarity
scores~$s_j(\mathbf{x})$ but on a probability distribution over
candidates.
We obtain this by applying a temperature-scaled softmax,
\begin{equation}\label{eq:candidate_prob}
  p_j(\mathbf{x})
  = \frac{\exp\!\bigl[s_j(\mathbf{x})/T\bigr]}
         {\sum_{k=1}^{M_i} \exp\!\bigl[s_k(\mathbf{x})/T\bigr]}\,,
  \qquad j = 1, \ldots, M_i\,,
\end{equation}
with temperature $T > 0$.
The resulting distribution converts raw
similarity scores into a categorical distribution over candidates,
which serves as the basis for both the first-order and the
uncertainty-decomposition-based scoring functions below.

\bmhead{First-order uncertainty scores}
The simplest scoring functions require only a single predicted
candidate distribution and no access to an ensemble or posterior
approximation.
The \emph{confidence} $\kappa_{\mathrm{conf}} = \max_j\, p_j(\mathbf{x})$
equals the probability assigned to the top-ranked candidate,
following the maximum softmax probability
baseline~\cite{hendrycks2017baseline} widely used for
misclassification detection.
The \emph{score gap}
$\kappa_{\mathrm{gap}} = s_{(1)}(\mathbf{x}) - s_{(2)}(\mathbf{x})$
measures the difference between the two highest pre-softmax similarity
scores, indicating how separated the top candidate is
from its nearest competitor.
Both scores capture how peaked the candidate distribution is.
Because they require no repeated inference, they are
computationally inexpensive and applicable to any retrieval model.

\begin{table*}[t]
\caption{Overview of different scoring functions~$\kappa$ investigated in this
work.  All scores are oriented so that higher values indicate
greater confidence.  ``Order'' indicates whether the score requires
a single prediction (1st) or multiple
samples from a posterior distribution (2nd).  Formal definitions are given in
Appendix~\ref{appendix: uncertainty metrics}.}
\label{tab:scoring-functions}
\centering
\footnotesize
\setlength{\tabcolsep}{4pt}
\renewcommand{\arraystretch}{1.05}
\begin{tabular*}{\textwidth}{@{\extracolsep{\fill}}l l l p{0.43\textwidth}@{}}
\toprule
Score & Level & Order & Description \\
\midrule
$\kappa_{\mathrm{conf}}$ & Retrieval & 1st
  & Max.\ candidate probability \\
$\kappa_{\mathrm{gap}}$  & Retrieval & 1st
  & Difference between top-two similarity scores \\
\addlinespace
$\kappa_{\mathrm{bit}}^{\mathrm{tot}}$  & Fingerprint & 2nd
  & Total predictive entropy (fingerprint) \\
$\kappa_{\mathrm{bit}}^{\mathrm{al}}$   & Fingerprint & 2nd
  & Aleatoric uncertainty (fingerprint) \\
$\kappa_{\mathrm{bit}}^{\mathrm{ep}}$   & Fingerprint & 2nd
  & Epistemic uncertainty (fingerprint) \\
\addlinespace
$\kappa_{\mathrm{ret}}^{\mathrm{tot}}$  & Retrieval & 2nd
  & Total predictive entropy (candidate ranking) \\
$\kappa_{\mathrm{ret}}^{\mathrm{al}}$   & Retrieval & 2nd
  & Aleatoric uncertainty (candidate ranking) \\
$\kappa_{\mathrm{ret}}^{\mathrm{ep}}$   & Retrieval & 2nd
  & Epistemic uncertainty (candidate ranking) \\
$\kappa_{\mathrm{rank}}$ & Retrieval & 2nd
  & Rank variance of top-$K$ set across samples \\
\addlinespace
$\kappa_{\mathrm{knn}}$  & Input & 1st
  & Deep $k$-nearest-neighbor distance \\
$\kappa_{\mathrm{mah}}$  & Input & 1st
  & Mahalanobis distance \\
\bottomrule
\end{tabular*}
\end{table*}

\bmhead{Second-order uncertainty scores}
A richer family of scoring functions decomposes predictive uncertainty
into aleatoric and epistemic components.
This decomposition requires not a single prediction, but a distribution
over predictions, obtained by placing a \emph{second-order distribution}
\rev{$q(\boldsymbol{\theta}\mid\mathbf{x})$} over the model's predicted
fingerprint probabilities \cite{bengs2022pitfalls}.
In practice, this distribution is intractable and is approximated by
drawing $S$ samples, e.g.\ from independently trained ensemble
members or stochastic forward passes.
\rev{We index these samples by parenthesized superscripts,
$\boldsymbol{\theta}^{(r)}$, $r=1,\ldots,S$. The sample-specific candidate
score is
$s_j^{(r)}(\mathbf{x})
= \mathrm{sim}(\boldsymbol{\theta}^{(r)},\mathbf{c}_j)$, and
$p_j^{(r)}(\mathbf{x})$ denotes the corresponding candidate probability
obtained through Eqn.~\eqref{eq:candidate_prob}.}
Given these samples, the information-theoretic
decomposition~\cite{depeweg2018decomposition} then decomposes the
entropy of the averaged prediction (total uncertainty) into the mean entropy of
individual sample predictions (aleatoric uncertainty) and their mutual
information with the model parameters (epistemic uncertainty).

\bmhead{Uncertainty levels} The decomposition can be applied at two levels.
At the \emph{fingerprint level}, each predicted bit probability
$\theta_d$ is treated as a Bernoulli parameter, and the decomposition
is applied independently to each of the $D$ bits and then
aggregated~\cite{sale2024labelwisealeatoricepistemicuncertainty},
yielding scores
$\kappa_{\mathrm{bit}}^{\mathrm{tot}}$,
$\kappa_{\mathrm{bit}}^{\mathrm{al}}$, and
$\kappa_{\mathrm{bit}}^{\mathrm{ep}}$.
At the \emph{retrieval level}, the decomposition is instead applied
to the candidate probability distributions induced by each posterior
sample, yielding retrieval-level scores
$\kappa_{\mathrm{ret}}^{\mathrm{tot}}$,
$\kappa_{\mathrm{ret}}^{\mathrm{al}}$, and
$\kappa_{\mathrm{ret}}^{\mathrm{ep}}$.
These scores capture uncertainty about which candidate is
correct rather than about individual fingerprint bits.
Which level provides the more informative selection criterion is an
empirical question addressed in Section~\ref{sec: results and discussion}.
As an additional second-order measure, we consider the \emph{rank variance}
$\kappa_{\mathrm{rank}}$, which quantifies the stability of the top-$K$
candidate set across posterior samples.
For each input, the top-$K$ candidates are identified under the mean
prediction and re-ranked under every posterior sample, and the rank variance
is the average variance of these ranks across samples.
The motivation is structural: for $K=1$, retrieval success reduces
to identifying the top candidate, which confidence directly captures.
For $K > 1$, it becomes a set-membership question, and
the relevant uncertainty concerns the stability of the predicted set's composition.
Because it captures both closely spaced similarity scores and model disagreement,
 rank variance acts as an implicit measure of total predictive uncertainty at
the ranking level \cite{adomavicius2016classification}.

\bmhead{Distance-based scores}
The measures above quantify uncertainty via the model's predictions.
An orthogonal approach assesses whether the input itself lies in a
region of the learned representation space that is well covered by the
training data~\cite{yang2024generalized}.
We consider the \emph{deep $k$-nearest-neighbor distance}
$\kappa_{\mathrm{knn}}$~\cite{sun2022out}, which measures the average
Euclidean distance of the input's penultimate-layer representation
to its $k$ nearest training neighbors,
and the \emph{Mahalanobis distance}
$\kappa_{\mathrm{mah}}$~\cite{lee2018simple}, which accounts for the
covariance structure of the training representations.
The rationale is that spectra far from the training distribution
are more likely to be poorly predicted.

\subsection{Risk control with statistical guarantees}
\label{sec:risk-control}

The scoring functions of Section~\ref{sec:scoring-functions} rank inputs by
confidence, but do not prescribe a specific threshold~$\tau$.
In practice, a practitioner requires a concrete decision rule:
Given a tolerable target risk rate $r^*$, which predictions should be accepted?
Setting the threshold on a held-out set to match the target risk empirically does
not account for finite-sample variability and can lead to violations of the
target risk on new data \cite{vapnik1999overview}.
We therefore adopt the \emph{selection with guaranteed risk}~(SGR)
algorithm~\cite{geifman2017selective}, \rev{a procedure that chooses~$\tau$ with provable guarantees
on the selective risk.}
SGR finds a threshold~$\tau^*$ such that
\begin{equation}\label{eq:sgr_guarantee}
  \mathbb{P}\bigl(R(f, g_{\tau^*}) > r^*\bigr) < \delta\,,
\end{equation}
for user-specified target risk~$r^* > 0$ and confidence
level~$0 < \delta < 1$.
Using a held-out calibration set, the algorithm computes \rev{finite-sample}
binomial tail bounds on the true selective risk at each candidate
threshold and applies a union bound to preserve the overall
confidence level~$\delta$.
Among all thresholds satisfying the bound, SGR selects the one that
maximizes coverage.
Full algorithmic details are given in
Appendix~\ref{appendix:risk-control}.
\rev{For our experiments, we split
the test dataset into equally sized calibration and evaluation halves and set
$\delta = 0.001$.}
The threshold~$\tau^*$ is selected on the calibration half,
while the reported coverages and empirical risks are measured on the evaluation half.
\rev{Conditional on the data being exchangeable, SGR calibrates the threshold on held-out data with
a distribution-free finite-sample bound on the chosen risk, requiring neither a decoy model nor calibrated
posteriors.}

\subsection{Evaluation}
\label{sec:evaluation}

We evaluate the quality of different scoring functions for selective prediction twofold: by evaluating the area under
the risk-coverage curve, and assessing the coverage of
the risk-controlled selective classifier.

\bmhead{Risk-coverage curves}
For a given scoring function~$\kappa$ which defines the selector $g_{\tau}$, the
\emph{risk-coverage curve} is the parametric curve
$\tau \mapsto \bigl(\varphi(f, g_\tau),\; R(f, g_\tau)\bigr)$
obtained by varying the threshold~$\tau$ over its full range.
As $\tau$ increases, coverage decreases and, for a well-chosen
$\kappa$, the selective risk decreases with it.
The curve thus visualizes the trade-off between committing to
more predictions and controlling the error rate among them.
Its quality is summarized by the
\emph{area under the risk-coverage curve},
\begin{equation}\label{eq:aurc}
  \mathrm{AURC}(\kappa)
  = \int_0^1 R(\varphi)\, \mathrm{d}\varphi\,,
\end{equation}
where $R(\varphi)$ denotes the selective risk at coverage
level~$\varphi$, i.e.\ the value of
$R(f, g_\tau)$ at the unique threshold~$\tau$ satisfying
$\varphi(f,g_\tau) = \varphi$.
Lower values indicate a more favourable risk-coverage
trade-off. 

The AURC of a scoring function is determined by two
factors: the base error rate of the model and the ranking
quality of the scoring function $\kappa$.
A model with high overall error will have high AURC even if
$\kappa$ perfectly separates correct from incorrect
predictions.
Since this paper is concerned with the quality of the scoring
function rather than the absolute performance of the
underlying model, we need a metric that isolates the
contribution of it from the base error rate.
To this end, we compare against two reference
values.
The \emph{oracle}
$\mathrm{AURC}_{\mathrm{oracle}}$ is obtained by
sorting instances by decreasing loss, thereby rejecting all
incorrect predictions first.
It represents the lowest achievable AURC for the given model
and loss.
The \emph{random} baseline
$\mathrm{AURC}_{\mathrm{random}}$ rejects instances in
arbitrary order, yielding a constant risk equal to the
overall error rate at every coverage level.
We report the \emph{relative AURC},
\begin{equation}\label{eq:rel_aurc}
  \mathrm{relAURC}(\kappa)
  = \frac{\mathrm{AURC}(\kappa)
          - \mathrm{AURC}_{\mathrm{oracle}}}
         {\mathrm{AURC}_{\mathrm{random}}
          - \mathrm{AURC}_{\mathrm{oracle}}}\,,
\end{equation}
which normalizes the excess over the oracle by the
achievable range of improvement, giving
$\mathrm{relAURC} = 0$ for a perfect scoring function and
$\mathrm{relAURC} = 1$ for random rejection.

\bmhead{Coverage at target risk}
For the risk-controlled setting, we report the \emph{coverage at
target risk}: given a target risk rate $r^*$ and a confidence
parameter $\delta>0$, the SGR algorithm
returns an optimal threshold $\tau^*$ used for deciding 
whether to accept or reject each prediction.
We report the empirical selective risk at this threshold on the held-out evaluation data,
and the
corresponding coverage
\begin{equation*}
  \varphi(f, g_{\tau^*})
   = \mathbb{E}\!\left[g_{\tau^*}(\mathbf{x})\right]
   = \mathbb{P}\!\left[\kappa(\mathbf{x}) \geq \tau^*\right],
\end{equation*}
whose empirical estimate is the fraction of instances for which the model commits to a
prediction while satisfying the risk constraint.
Higher coverage at the same target risk indicates a more informative
scoring function.
We evaluate this metric over a range of target risks
$r^*$, while also verifying that the empirical
risk on the evaluation data respects the desired bound.

\subsection{Experimental setup}
\label{sec:experimental-setup}
\bmhead{Dataset} We use the MassSpecGym benchmark~\cite{bushuiev2024massspecgym},
which consists of $231\,104$ spectra-molecule pairs covering $28\,929$
unique molecules.
\rev{The benchmark provides a predefined split into} into training, validation and test sets by
clustering on maximum common edge subgraph (MCES)
distance~\cite{raymond2002maximum},
ensuring that no molecular structure appears in more than one split.
\rev{A minimum MCES distance of~$10$ is enforced between molecules assigned to different
folds, the split is structure-disjoint at the level of molecular structures rather than of individual
spectra.}
This yields $194\,119$, $19\,429$, and $17\,556$ spectra
covering $22\,746$, $3\,185$, and $2\,998$
molecules, respectively.
Each molecule is represented by a $4096$-dimensional Morgan
fingerprint~\cite{rogers2010extended} computed with radius~$2$.
For each spectrum, a candidate set $\mathcal{C}_i$ is constructed 
by iteratively filtering molecules with matching molecular formula from
three chemical databases of increasing size, until the cap of
$|\mathcal{C}_i| \leq 256$ is reached~\cite{bushuiev2024massspecgym}.
The resulting candidate set sizes vary considerably across
instances and directly affect task difficulty. This variation is analyzed
further in Section~\ref{sec:candidate-set-analysis} and Appendix~\ref{sec: dataset characterisation}.
\rev{In untargeted metabolomics, formula-based candidate sets can substantially exceed this cap and
the molecular formula may itself be uncertain. Appendix~\ref{appendix:candidate-sets} therefore analyses two different scenarios: 
an uncapped candidate set with candidates filtered by molecular formula, and a precursor-mass-filtered candidate set.}

\bmhead{Model and training procedure}
As base model we adopt the fingerprint predictor and training procedure of~\citet{de2026small}.\footnote{Our implementation is available at \url{https://github.com/mkjuergens/Selective-MSMS}.}
The input spectrum is binned and mapped by a three-layer fully-connected neural network 
to fingerprint probabilities
$\boldsymbol{\theta} = \sigma(f(\mathbf{x})) \in [0,1]^{4096}$.
Architectural details are given in Appendix \ref{sec: model-architecture}.
While more expressive architectures exist~\cite{goldman2023annotating},
the multi-layer perceptron (MLP) achieves competitive retrieval performance on
MassSpecGym~\cite{de2026small}, and its standard feedforward
structure permits clean application of established posterior
approximation methods
without architectural modifications.
\rev{To verify that our findings are not specific to this architecture, we additionally evaluate a
transformer spectrum encoder trained with the same ranking loss. Its architecture and main results are
given in Appendix~\ref{appendix:transformer}.}
The model is trained with a contrastive ranking loss
that directly optimizes retrieval performance.
For each training instance $(\mathbf{x}_i, \mathbf{y}_i, \mathcal{C}_i)$,
the loss is the negative log-likelihood of the true candidate
$\mathbf{y}_i$ under the temperature-scaled softmax over the
candidate set,
\begin{equation}\label{eq:ranking_loss}
  \ell_{\mathrm{rank}}(\mathbf{x}_i, \mathbf{y}_i)
  = -\log
    \frac{\exp\!\bigl[s_{\boldsymbol{y}_i}(\mathbf{x}_i)/T\bigr]}
         {\sum_{j=1}^{M_i}\exp\!\bigl[s_j(\mathbf{x}_i)/T\bigr]}\,,
\end{equation}
where $s_j(\mathbf{x})$ denotes the cosine similarity between the predicted probability vector
as defined in
Eqn.~\eqref{eq:cosine_sim}, and $T$ is a temperature hyperparameter.
\rev{The temperature is selected according to De Waele et al.~\cite{de2026small} as
$T = 0.003$ for the ranking loss.}
We additionally train models with focal loss~\cite{lin2017focal}
on individual bits for the analysis of fingerprint-level
uncertainty. In this setting, the predicted fingerprint retains its
interpretation as a vector of per-bit probabilities rather than
serving primarily as a retrieval embedding.

\bmhead{Uncertainty estimation}
The fingerprint- and retrieval-level scoring functions
described in Section~\ref{sec:scoring-functions} require a second-order
distribution \rev{$q(\boldsymbol{\theta}\mid\mathbf{x})$},
which allows for the disentanglement of uncertainty estimates into aleatoric and epistemic components.
Our primary second-order method, whose results are shown throughout the main text, is a
Deep Ensemble~\cite{lakshminarayanan2017simple} consisting of
$S = 5$ copies of the full model, trained independently with
different random initializations and data-loader shuffles.
We additionally evaluate MC~Dropout~\cite{gal2016dropout} using $S=50$ stochastic
forward passes, and the last-layer Laplace
approximation~\cite{daxberger2021laplace} with $S=50$ weight samples.
For the distance-based scores $\kappa_{\mathrm{knn}}$ and
$\kappa_{\mathrm{mah}}$, we extract the $1024$-dimensional
penultimate-layer representations from a single ensemble member
for all training and test spectra.
Representations are $\ell_2$-normalized before distance computation.
For $\kappa_{\mathrm{knn}}$, we use $k=100$ neighbors.
\rev{\bmhead{Supervised score combination}
Instead of using one single score for selection, one could also use a combination of multiple scores.
To test this idea, we fit a separate logistic
regression model on different uncertainty scores to predict hit rate, where the model is fit on the MassSpecGym validation fold. 
This experiment follows COSMIC's supervised-combination principle but
is not an implementation of it, since CSI:FingerID and fragmentation-tree features are unavailable
in our pipeline. Full details and results are given in Appendix~\ref{appendix:supervised-baseline}.}

\bmhead{Evaluation losses}
All primary analyses use the retrieval loss
$\ell_K = 1 - \mathrm{Hit@}K$ for $K \in \{1, 5, 20\}$.
To test whether the alignment between scoring function
and task loss extends bidirectionally, we additionally
evaluate with fingerprint-level similarity losses
$\ell_{\mathrm{sim}} = 1 - \mathrm{sim}(\hat{\mathbf{y}},
\mathbf{y})$ using Tanimoto, cosine, and Hamming loss as similarity measures. The results are given in
Appendix~\ref{sec: similarity loss}.

\bmhead{Prediction aggregation}

For each input, the $S$ posterior samples must be aggregated into a single
candidate ranking. We consider three aggregation levels:
\begin{enumerate}
  \item \emph{Fingerprint-level aggregation:} average the predicted
  fingerprints,
  $\overline{\boldsymbol{\theta}}
  = \frac{1}{S}\sum_{r=1}^{S}\boldsymbol{\theta}^{(r)}$,
  and rank candidates by
  $\mathrm{sim}(\overline{\boldsymbol{\theta}},\mathbf{c}_j)$.
  \item \emph{Score-level aggregation:} rank candidates by the mean score
  $\overline{s}_j(\mathbf{x})
  = \frac{1}{S}\sum_{r=1}^{S}s_j^{(r)}(\mathbf{x})$.
  \item \emph{Probability-level aggregation:} rank candidates by the mean
  candidate probability
  $\overline{p}_j(\mathbf{x})
  = \frac{1}{S}\sum_{r=1}^{S}p_j^{(r)}(\mathbf{x})$.
\end{enumerate}
The three aggregation strategies yield overall comparable retrieval performance. In the following, we adopt
score-level aggregation when analyzing retrieval performance, and fingerprint-level aggregation
when analyzing fingerprint similarity.

\section{Results and discussion}
\begin{figure*}[t]
  \centering
  \includegraphics[width=\textwidth]{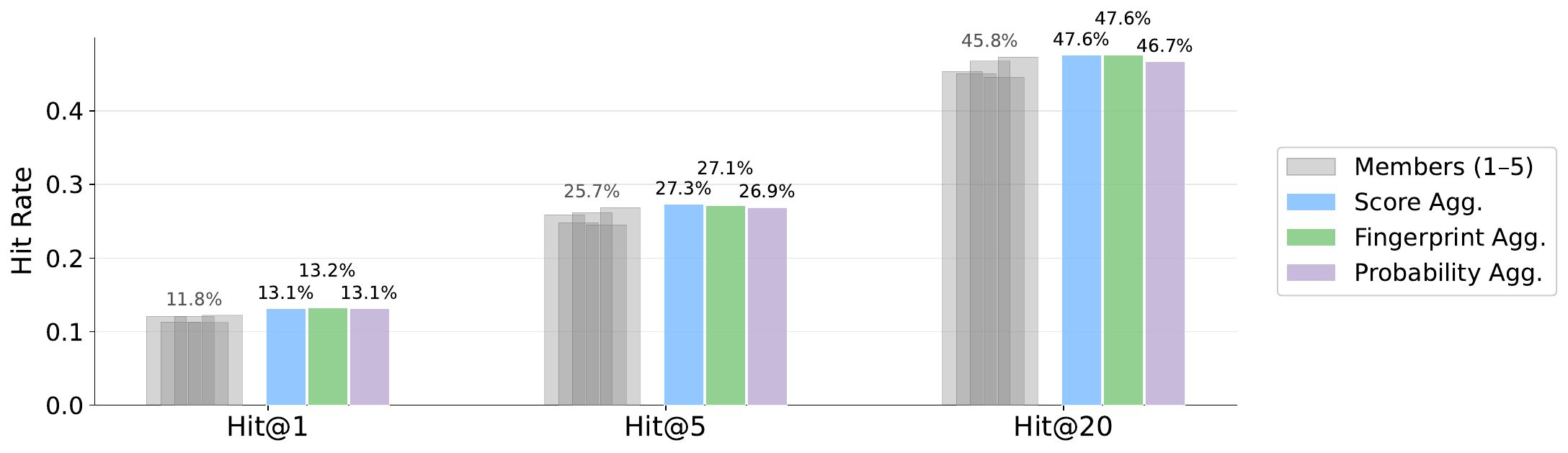}
\caption{Average $\mathrm{Hit}@K$ rate for $K\in\{1, 5, 20\}$ for samples from the second-order distribution (grey)
and their aggregate (blue, green, purple). Results are shown for a Deep Ensemble of $S=5$ members and the experimental setup
as described in Section \ref{sec:experimental-setup}, evaluated on the test set of MassSpecGym with candidates filtered by molecular formula. 
The aggregate is computed using
 the different aggregation strategies described in Section \ref{sec:experimental-setup}.}
 \label{fig: aggregation}
\end{figure*}
\label{sec: results and discussion}
\rev{Unless stated otherwise, experiments use formula-filtered MassSpecGym test
candidate sets and} the
evaluation metrics and experimental setup as defined in Section~\ref{sec:evaluation}.
We first briefly analyze the baseline retrieval performance of the different
uncertainty estimation methods and the effect of prediction
aggregation.
We then compare the effect of different scoring functions via a risk-coverage analysis,
and subsequently examine how the candidate set size influences selective prediction
quality.
Finally, we assess the coverage attainable under risk guarantees for different target risks.

\subsection{Baseline retrieval performance}
\label{sec:baseline}
Table~\ref{tab:baseline} reports the retrieval performance of the aggregate of the three
different uncertainty estimation methods on the MassSpecGym test set.
For both training objectives, aggregating over multiple posterior
samples consistently improves hit rates over the average individual prediction (Avg.\ sample),
with the Deep Ensemble leading to the biggest improvement.
Across the three second-order methods,
the aggregate of the Deep Ensemble trained with ranking loss leads to the highest hit rates. 
Figure~\ref{fig: aggregation} illustrates the per-member variability
and the effect of aggregation. Individual members exhibit considerable variance in hit rate, and
their predictions disagree on a substantial fraction of test
instances.

\begin{table*}[t]
\caption{Retrieval performance on the MassSpecGym test set for the \rev{two different model architectures} discussed in Section~\ref{sec:experimental-setup}, under different second-order uncertainty estimation methods.
$\mathrm{Hit@}K$ (\%) denotes the fraction of test spectra for which the
correct molecule appears among the top-$K$ retrieved candidates.
$S$ denotes the number of posterior samples (ensemble members or
stochastic forward passes).
Aggregate: prediction obtained by averaging the $S$ predicted
scores before ranking.
Avg.\ sample: mean $\mathrm{Hit@}K$ across the $S$
individual predictions.}
\label{tab:baseline}
\centering
\footnotesize
\setlength{\tabcolsep}{4pt}
\renewcommand{\arraystretch}{1.05}
\begin{tabular*}{\textwidth}{@{\extracolsep{\fill}}ll rrr rrr@{}}
\toprule
 & & \multicolumn{3}{c}{Aggregate}
   & \multicolumn{3}{c}{Avg.\ sample} \\
\cmidrule(lr){3-5}\cmidrule(lr){6-8}
Training loss & Method & Hit@1 & Hit@5 & Hit@20 & Hit@1 & Hit@5 & Hit@20 \\
\midrule
\multicolumn{8}{@{}l}{\rev{\textit{MLP}}} \\
\multirow{3}{*}{Focal}
  & \textbf{Deep Ensemble} ($S\!=\!5$)   & \textbf{11.43} & \textbf{23.80} & \textbf{42.36} & 10.71 & 22.67 & 40.70 \\
  & MC Dropout ($S\!=\!50$)     & 11.13 & 22.78 & 40.75 & 10.33 & 22.10  & 40.15  \\
  & Laplace ($S\!=\!50$)        &  11.22 & 22.62 & 41.39 & 11.01 & 22.65 & 41.38 \\
\midrule
\multirow{3}{*}{Ranking}
  & \textbf{Deep Ensemble} ($S\!=\!5$)   & \textbf{13.12} & \textbf{27.29} & \textbf{47.57} & 11.83 & 25.66 & 45.83 \\
  & MC Dropout ($S\!=\!50$)     & \rev{11.67} & \rev{25.86} & \rev{45.41} & 11.49 & 25.32 & 44.67 \\
  & Laplace ($S\!=\!50$)        & \rev{12.22} & \rev{26.83} & \rev{47.25} & 12.11 & 26.35 & 46.81 \\
\midrule
\multicolumn{8}{@{}l}{\rev{\textit{Transformer}}} \\
\rev{Ranking}
  & \rev{\textbf{Deep Ensemble} ($S\!=\!5$)}   & \rev{\textbf{17.44}} & \rev{\textbf{35.64}} & \rev{\textbf{57.02}} & \rev{14.77} & \rev{31.28} & \rev{52.69} \\
\bottomrule
\end{tabular*}
\end{table*}

\subsection{Risk-coverage curves}
\label{sec: risk-coverage analysis}
\begin{figure*}[t]
  \centering
  \includegraphics[width=\textwidth]{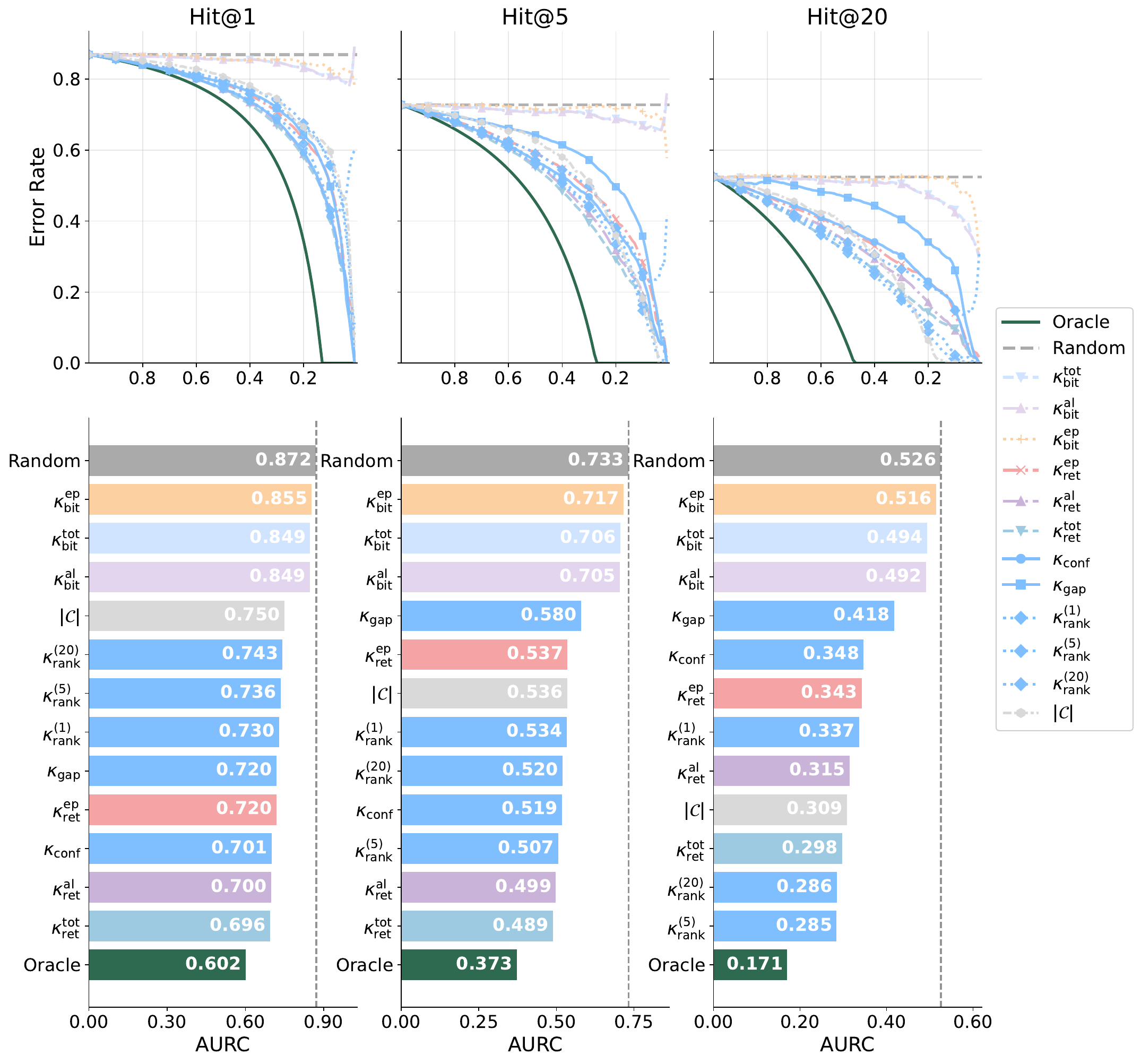}

\caption{Risk-coverage analysis for different scoring functions $\kappa$ based on different uncertainty estimates 
as described in Section \ref{sec:scoring-functions}.
The selective risk is calculated using $\ell_K = 1 - \mathrm{Hit@}K$ for $K \in \{1, 5, 20\}$.
Results are shown for a Deep Ensemble of $S=5$ members and the MLP model trained with ranking loss,
as described in Section \ref{sec:experimental-setup}, evaluated on the test set of MassSpecGym
 with candidates filtered by molecular formula.
Colors encode the uncertainty component:
  blue tones indicate total uncertainty,
  purple indicates aleatoric uncertainty,
  and red tones indicate epistemic uncertainty.
  Darker shades correspond to retrieval-level scores,
  lighter shades to fingerprint-level scores.
\textbf{(a)}~Risk-coverage curves. \textbf{(b)}~AURC values.}
\label{fig: risk-coverage deep ensemble}
\end{figure*}

To evaluate how effectively each scoring function reduces the selective risk, we use the metrics
described in
Section~\ref{sec:evaluation},
with the retrieval loss $\ell_K = 1 - \mathrm{Hit@}K$, for $K\in\{1,5,20\}$.
\rev{Unless stated otherwise, all softmax-derived scores use the fixed evaluation
temperature $T_{\mathrm{eval}}=0.003$, equal to the temperature used for
training. The sensitivity to this choice is examined in
Appendix~\ref{appendix:temperature}.}
Figure~\ref{fig: risk-coverage deep ensemble} shows the obtained risk-coverage curves and the corresponding AURC values
for the MLP Deep Ensemble model trained with ranking loss. Table~\ref{tab:aurc} 
reports the corresponding relative AURC values across all three second-order methods.
\begin{table*}[t]
\caption{%
  Selective prediction quality for different scoring functions,
  evaluated by relAURC. Lower is better.
  Results are shown for different second-order uncertainty estimation methods and the MLP model trained with ranking loss, evaluated on the test set of MassSpecGym with candidates filtered by molecular formula.
  The selective risk is evaluated using hit rate, $\ell_K = 1 - \mathrm{Hit}@K$, $K=1,5,20$.
  relAURC $= 0$ corresponds to perfect (oracle) rejection,
  relAURC $= 1$ corresponds to random rejection.}
\label{tab:aurc}
\centering
\footnotesize
\setlength{\tabcolsep}{2pt}
\renewcommand{\arraystretch}{1.05}
\color{blue}
\begin{tabular*}{\textwidth}{@{\extracolsep{\fill}}l rrr rrr rrr@{}}
\toprule
 & \multicolumn{3}{c}{Deep Ensemble}
 & \multicolumn{3}{c}{MC Dropout}
 & \multicolumn{3}{c}{Laplace} \\
\cmidrule(lr){2-4}\cmidrule(lr){5-7}\cmidrule(lr){8-10}
$\kappa$
 & $\ell_1$ & $\ell_5$ & $\ell_{20}$
 & $\ell_1$ & $\ell_5$ & $\ell_{20}$
 & $\ell_1$ & $\ell_5$ & $\ell_{20}$ \\
\midrule
\multicolumn{10}{@{}l}{\textit{Retrieval-level}} \\
\quad $\kappa_{\mathrm{conf}}$        & 0.366 & 0.406 & 0.498 & 0.397 & 0.440 & 0.527 & 0.419 & 0.521 & 0.614 \\
\quad $\kappa_{\mathrm{gap}}$         & 0.439 & 0.576 & 0.696 & 0.461 & 0.591 & 0.702 & 0.465 & 0.627 & 0.733 \\
\quad $\kappa_{\mathrm{ret}}^{\mathrm{tot}}$
  & \textbf{0.347} & \textbf{0.323} & 0.359
  & \textbf{0.369} & \textbf{0.358} & 0.388
  & \textbf{0.392} & 0.442 & 0.500 \\
\quad $\kappa_{\mathrm{ret}}^{\mathrm{al}}$  & 0.364 & 0.351 & 0.407 & 0.373 & 0.366 & 0.407 & 0.394 & 0.446 & 0.508 \\
\quad $\kappa_{\mathrm{ret}}^{\mathrm{ep}}$  & 0.438 & 0.455 & 0.485 & 0.464 & 0.505 & 0.516 & 0.461 & 0.511 & 0.522 \\
\quad $\kappa_{\mathrm{rank}}$\; $K\!=\!1$         & 0.473 & 0.448 & 0.469 & 0.505 & 0.469 & 0.516 & 0.713 & 0.686 & 0.729 \\
\quad $\kappa_{\mathrm{rank}}$\; $K\!=\!5$         & 0.498 & 0.373 & \textbf{0.322} & 0.495 & 0.379 & 0.329 & 0.533 & 0.418 & 0.428 \\
\quad $\kappa_{\mathrm{rank}}$\; $K\!=\!20$        & 0.524 & 0.410 & 0.324 & 0.496 & 0.397 & \textbf{0.327} & 0.528 & \textbf{0.404} & \textbf{0.345} \\
\midrule
\multicolumn{10}{@{}l}{\textit{Fingerprint-level}} \\
\quad $\kappa_{\mathrm{bit}}^{\mathrm{tot}}$       & 0.917 & 0.927 & 0.911 & 0.895 & 0.903 & 0.901 & 0.944 & 0.935 & 0.928 \\
\quad $\kappa_{\mathrm{bit}}^{\mathrm{al}}$        & 0.916 & 0.924 & 0.905 & 0.895 & 0.903 & 0.900 & 0.944 & 0.934 & 0.927 \\
\quad $\kappa_{\mathrm{bit}}^{\mathrm{ep}}$        & 0.939 & 0.957 & 0.971 & 0.920 & 0.900 & 0.904 & 0.960 & 0.970 & 0.962 \\
\midrule
\multicolumn{10}{@{}l}{\textit{Input-level}\textsuperscript{\dag}} \\
\quad $\kappa_{\mathrm{knn}}$
  & 0.985 & 0.915 & 0.929
  & ---   & ---   & ---
  & ---   & ---   & --- \\
\quad $\kappa_{\mathrm{mah}}$
  & 0.977 & 0.928 & 0.893
  & ---   & ---   & ---
  & ---   & ---   & --- \\
\midrule
\multicolumn{10}{@{}l}{\textit{Other}} \\
\quad $|\mathcal{C}|$
  & 0.548 & 0.453 & 0.390
  & 0.509 & 0.421 & 0.370
  & 0.531 & 0.432 & 0.373 \\
\bottomrule
\end{tabular*}
\par\smallskip{\footnotesize
  \textsuperscript{\dag}Distance-based scores depend only on the
  encoder representation, not the posterior approximation method.
  Values shown are for one Deep Ensemble member. See Appendix \ref{appendix:distance}.\par}
\end{table*}
Interestingly, the best scoring function depends on the size of the selected set $\mathcal{S}_K$ for which the hit rate 
is evaluated.
For $K\!=\!1$ and $K\!=\!5$, retrieval-level total uncertainty achieves the lowest
relAURC for the Deep Ensemble.
For $K\!=\!20$, however, the rank variance becomes the strongest criterion,
although total and aleatoric retrieval-level uncertainty remain competitive.
This change
is consistent with the structure of the retrieval loss. Hit rate for $K > 1$
is no longer a question about the candidate with highest probability alone, but about
set membership, i.e.\ whether the correct candidate
falls within the top-$K$ set. \rev{The candidate-set-size criterion shows a
similar improvement at broader cutoffs, and its correlation with the
cutoff-matched rank variance increases from $\rho=0.32$ for $K=1$ to
$\rho=0.78$ for $K=20$ (Appendix~\ref{sec:correlation-analysis}). Rank
variance nevertheless remains the stronger criterion.}

The same broad pattern is observed for the other posterior approximations.
Total retrieval uncertainty is strongest for $K\!=\!1$ with MC Dropout and
Laplace, and for $K\!=\!5$ with MC Dropout. Rank variance is strongest for
$K\!=\!20$ with both methods and also for $K\!=\!5$ with Laplace. Rank variance
therefore provides a useful cutoff-specific measure of ranking instability,
particularly for broader MLP retrieval cutoffs, but is not uniformly the
strongest criterion.

The first-order scores provide computationally inexpensive alternatives
because they do not require posterior sampling. Confidence is the stronger
of these baselines and remains close to total uncertainty for $K\!=\!1$ across
the three posterior-approximation methods. The score gap also improves
substantially over random rejection, particularly for $K\!=\!1$, but becomes
less informative as the retrieval cutoff increases. Thus, confidence and
score gap provide practical alternatives when second-order uncertainty
estimation is unavailable, with confidence giving the more consistent
performance.

Across all three posterior approximations and retrieval cutoffs, total
retrieval uncertainty outperforms both its aleatoric and epistemic
components. Aleatoric uncertainty is generally close to total uncertainty,
whereas epistemic uncertainty alone is consistently weaker than both.
This suggests that disagreement between posterior samples captures only
part of the uncertainty relevant to retrieval failure. The result is
qualitatively consistent with the task-alignment principle of
\citet{hofman2025onesizedoesnotfitall}, stating that the optimal rejection criterion for selective prediction is 
total predictive uncertainty,  rather than the epistemic uncertainty component
alone.

In contrast, the fingerprint-level scores achieve relAURC values of
approximately $0.9$ or higher and are therefore close to random rejection.
We see that retrieval success depends not only on the absolute quality of the predicted
fingerprint, but on its similarity to the correct candidate \textit{relative} to
the competing candidates. A confidently predicted fingerprint may still
lead to incorrect retrieval when several candidates have similar
structures, whereas a less certain fingerprint can be sufficient when the
correct candidate is well separated. Fingerprint-level uncertainty is
therefore poorly aligned with the retrieval loss. Conversely, it becomes
informative when the target loss measures fingerprint similarity, as shown
in Appendix~\ref{sec: similarity loss}.

The input-level distance scores show similarly weak selective-prediction
performance. These scores measure whether a spectrum lies in a well-covered
region of the learned representation space, but this representation is
optimized for fingerprint prediction or retrieval and need not be organized
according to retrieval difficulty~\cite{li2025out}. Overall, the results
show that the level at which uncertainty is quantified is more important
than whether the score is conventionally interpreted as epistemic or
aleatoric.

\rev{To assess whether this ordering is specific to the MLP architecture,
Table~\ref{tab:transformer-relaurc-main} compares the MLP and transformer
Deep Ensembles. Both models use the same ranking objective and evaluation
setting. Full transformer results across all scoring functions are reported
in Appendix~\ref{appendix:transformer}.
\begin{table*}[t]
\caption{\rev{Robustness of the relAURC (Eqn.~\ref{eq:rel_aurc}, lower is better) ordering across architectures.
Values are given for the strongest retrieval-level scoring functions
for the MLP and transformer Deep Ensembles ($S=5$), on the MassSpecGym test set with candidates
filtered by molecular formula. Full transformer results across all scoring functions are given in Appendix~\ref{appendix:transformer}.}}
\label{tab:transformer-relaurc-main}
\centering
\footnotesize
\setlength{\tabcolsep}{4pt}
\renewcommand{\arraystretch}{1.05}
{\color{blue}
\begin{tabular*}{\textwidth}{@{\extracolsep{\fill}}l rrr rrr@{}}
\toprule
 & \multicolumn{3}{c}{MLP} & \multicolumn{3}{c}{Transformer} \\
\cmidrule(lr){2-4}\cmidrule(lr){5-7}
$\kappa$ & $\ell_1$ & $\ell_5$ & $\ell_{20}$ & $\ell_1$ & $\ell_5$ & $\ell_{20}$ \\
\midrule
$\kappa_{\mathrm{conf}}$\; confidence            & 0.366 & 0.406 & 0.498 & \textbf{0.356} & 0.367 & 0.420 \\
$\kappa_{\mathrm{gap}}$\; score gap              & 0.439 & 0.576 & 0.696 & 0.384 & 0.513 & 0.606 \\
$\kappa_{\mathrm{ret}}^{\mathrm{tot}}$\; total  & \textbf{0.347} & \textbf{0.323} & 0.359 & 0.365 & \textbf{0.317} & \textbf{0.320} \\
$\kappa_{\mathrm{ret}}^{\mathrm{al}}$\; aleatoric & 0.364 & 0.351 & 0.407 & 0.377 & 0.342 & 0.369 \\
$\kappa_{\mathrm{rank}}$\; $K\!=\!1$             & 0.473 & 0.448 & 0.469 & 0.446 & 0.404 & 0.401 \\
$\kappa_{\mathrm{rank}}$\; $K\!=\!5$             & 0.498 & 0.373 & \textbf{0.322} & 0.579 & 0.429 & 0.347 \\
$\kappa_{\mathrm{rank}}$\; $K\!=\!20$            & 0.524 & 0.410 & 0.324 & 0.609 & 0.500 & 0.409 \\
\bottomrule
\end{tabular*}}
\end{table*}
The architecture comparison reinforces the main result while qualifying
the role of rank variance. For the MLP, total uncertainty is strongest for
$K\,=\,1$ and $K\,=\,5$, while rank variance is strongest for $K\,=\,20$. For the
transformer, confidence is strongest for $K\,=\,1$ and total uncertainty is
strongest for $K\,=\,5$ and $K\,=\,20$. Rank variance remains informative for the
broader transformer cutoff, but does not outperform total uncertainty.
Thus, retrieval-level total uncertainty is the strongest criterion overall,
whereas the precise ordering of confidence and rank-stability scores
depends on the architecture and retrieval cutoff.}

\subsection{Candidate set size analysis}
\label{sec:candidate-set-analysis}
\begin{figure*}[t]
  \centering
  \includegraphics[width=\textwidth]{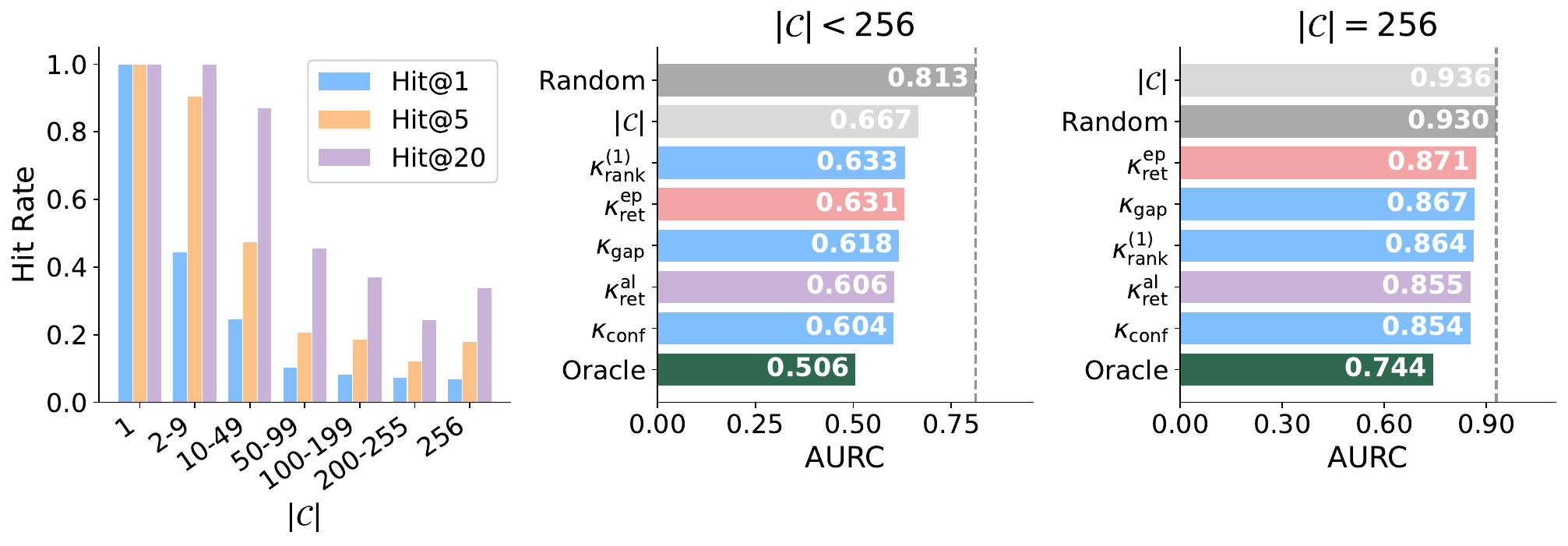}

\caption{Effect of the candidate set size on the retrieval performance and selective
  prediction quality. Results are shown for a Deep Ensemble of $S=5$ members and the experimental setup
as described in Section \ref{sec:experimental-setup}, evaluated on the test set of MassSpecGym with candidates
 filtered by molecular formula.
  \textbf{Left}: Average hit rate on subsets with binned candidate set size~$|\mathcal{C}|$.
  \textbf{Middle}, \textbf{Right}:~AURC values for $\ell_1 = 1 - \mathrm{Hit@}1$ and different scoring functions, on subsets of data 
  with $|\mathcal{C}| < 256$ vs. $|\mathcal{C}| = 256$.}
\label{fig:candidate-set-analysis}
\end{figure*}

A distinguishing feature of the retrieval setting is that the task difficulty varies across instances
 through the candidate
set size~$|\mathcal{C}|$, which ranges from one to the
benchmark cap of~$256$ and is determined by external
database queries.
Figure~\ref{fig:candidate-set-analysis}(a) confirms
the expected effect: hit rates decrease almost monotonically
with~$|\mathcal{C}|$.
To analyze the effect of candidate set size as a selection criterion, 
we stratify the test set into instances with varying number of candidates
($|\mathcal{C}| < 256$)
and instances with maximum number of candidates ($|\mathcal{C}| = 256$).
When candidate sets vary in size,
the gap between oracle and random rejection is wide and the scoring functions recover
a substantial fraction of the achievable improvement, as can be seen in Figure~\ref{fig:candidate-set-analysis} (b).
At the cap,
this gap shrinks and all scoring functions compress toward
the random baseline, reflecting the increased difficulty, see Figure~\ref{fig:candidate-set-analysis} (c).
The candidate set size itself, which is an effective
rejection criterion when it varies, becomes uninformative
once it is constant.
This illustration shows the role of the rank variance
$\kappa_{\mathrm{rank}}$ as a selection criterion.
\rev{Its association with candidate-set size increases with the retrieval
cutoff and is strongest for $K=20$. Its good performance at broader cutoffs
is therefore partly explained by this sensitivity:} larger sets produce
more candidates with similar scores near the decision boundary,
thereby inflating the rank variance.
However, the stratified view reveals that rank variance is
not merely a proxy for the number of candidates:
for the subset of instances with maximum number of candidates, rank variance still achieves
AURC values comparable to confidence and clearly below the random
baseline. This suggests that it captures both the task difficulty given by the number of candidates,
 and ranking instability in general.

\subsection{Coverage under risk control}
\label{sec:risk-control-results}

\begin{figure*}[t]
  \centering
  \includegraphics[width=\textwidth]{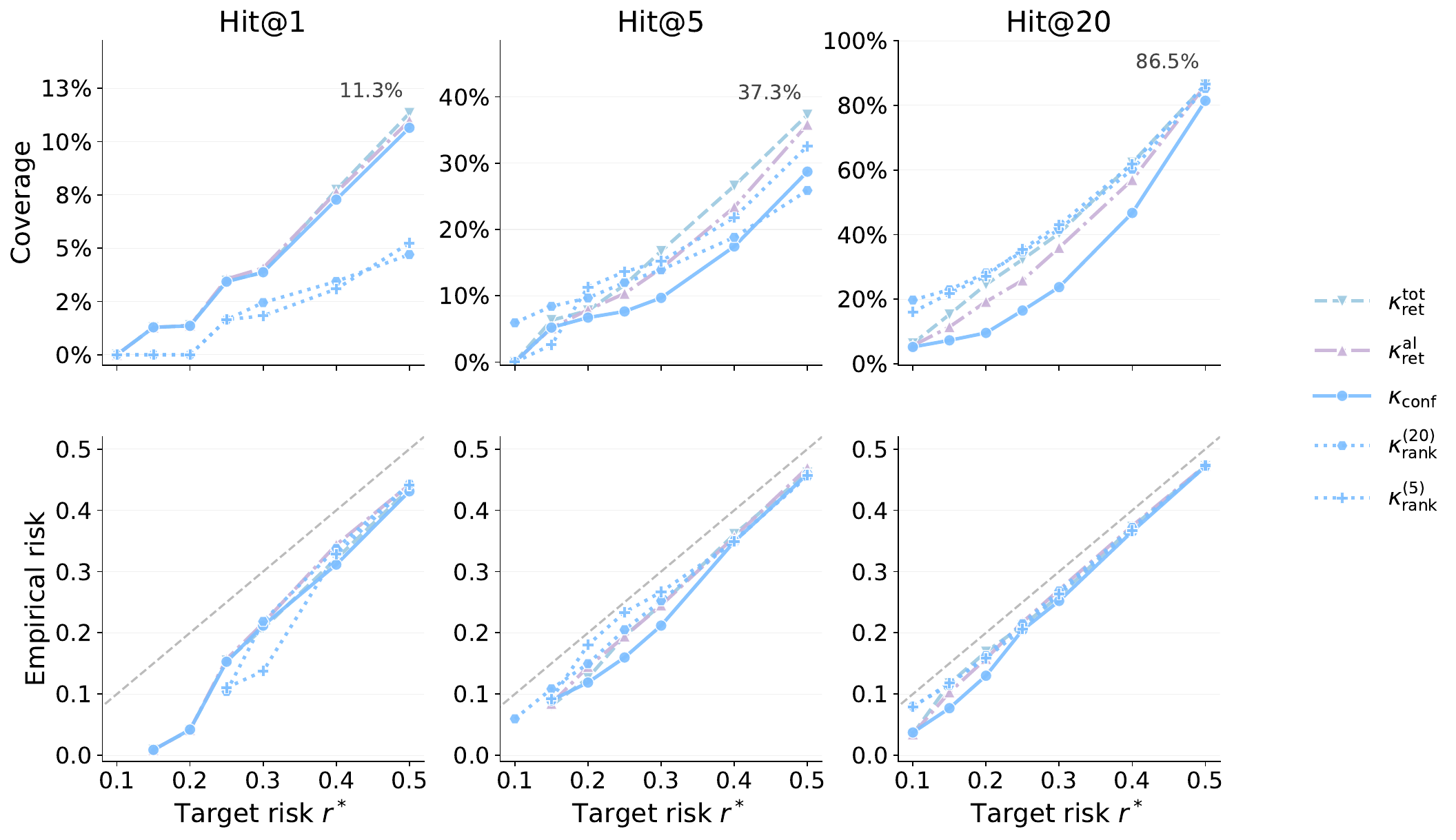}
  \caption{Risk-controlled annotation with the SGR algorithm with $\delta=0.001$ and different target risks $r^*$.
  Results are shown for the MLP Deep Ensemble of $S=5$ members and the experimental setup
as described in Section \ref{sec:experimental-setup}, with candidates filtered by molecular formula.
    The threshold~$\tau^*$ is calibrated on one half of the test set,
  coverage and empirical risk are evaluated on the held-out other half.
  \textbf{Top}: coverage attained at each target risk level~$r^*$
  for $K\in\{1,5,20\}$. Note the independent vertical scales.
  \textbf{Bottom}: empirical risk versus target risk. }
  \label{fig:sgr-coverage}
\end{figure*}

The previous sections evaluated scoring functions by their
ability to \emph{rank} predictions, asking which ones
correlate best with retrieval success.
We now turn to the question that matters most in practice:
\rev{given a tolerable error rate, how many molecule-level queries can be
annotated under the given risk guarantee?}

Figure~\ref{fig:sgr-coverage} (top row) shows the coverage
obtained by applying the SGR algorithm across a range of
target risk levels. 
We see that
the practical utility of risk-controlled selection depends
strongly on the task difficulty and the chosen risk level.
For controlling the hit rate at $K=20$, where the base error rate is moderate,
SGR retains up to 87\% of test spectra at a target risk
of~$r^* = 0.5$.
For $K=1$, however, the base error rate is high
and the algorithm can only guarantee risk control on a small subset of
the data.
This is not a limitation of the scoring functions but of
the task itself: when the majority of predictions are
incorrect, any procedure with valid risk guarantees must
\rev{abstain from making predictions for most instances}.
 In practice, this means that exact-match identification requires either stronger base models or an
acceptance of lower coverage, whereas candidate-list
settings already offer a practical operating
regime for automated annotation.
\rev{Total retrieval uncertainty and rank variance achieve the highest coverage
across the evaluated settings, consistent with the results in
Section~\ref{sec: risk-coverage analysis}. The corresponding transformer analysis is
reported in Appendix~\ref{appendix:transformer} and shows that a stronger baseline model also allows for higher coverage.}
\rev{Figure~\ref{fig:sgr-coverage} shows that the empirical risk on the held-out
evaluation half remains below the target at all evaluated points, thereby numerically confirming the risk control.}
\section{Conclusion}
\label{sec: conclusion}

We introduced a selective prediction framework for molecular structure
retrieval from tandem mass spectra and conducted a systematic evaluation of
uncertainty quantification strategies on the MassSpecGym benchmark.
By formulating retrieval as a selective classification problem with
task-specific losses, we evaluated scoring
functions from three complementary families, acting on the level of the molecular fingerprints,
the retrieval candidates, and the model's internal representation space.
We assessed their performance in terms of risk-coverage trade-offs,
\rev{and showed how generalization bounds can be used to obtain selected subsets with risk guarantees.}

\rev{Our analysis shows that the effectiveness of a scoring function depends
on how closely it reflects the target task. Retrieval-level scores provide
the most effective selection criteria for retrieval accuracy, with total
retrieval uncertainty performing best across most settings. Rank variance
remains competitive at larger retrieval cutoffs,
but its performance depends on the architecture and cutoff value. Confidence is
the strongest inexpensive baseline, while the score gap provides a useful
alternative for exact top-$1$ retrieval.}

This loss dependence extends to the granularity of
uncertainty estimation itself.
Fingerprint-level uncertainty, which quantifies how well the
model predicts individual molecular substructures, is
naturally aligned with fingerprint reconstruction quality but
proves a poor proxy for retrieval success, where what
matters is the \emph{relative} similarity between the
predicted fingerprint and the correct candidate versus its
competitors.
A spectrum whose fingerprint is confidently predicted may
still fail at retrieval when structurally similar candidates are
present, and conversely, a noisier prediction can succeed when
the correct candidate is well separated.
Epistemic uncertainty, whether at the fingerprint or retrieval
level, s consistently less effective than total or aleatoric uncertainty.
Distance-based scores likewise provide limited selective-prediction value.
\rev{Finally, we demonstrated that practitioners can specify a tolerable risk
and, under exchangeability, obtain a selected subset of annotations that
satisfies this constraint with high probability.}

Several limitations suggest directions for future research.
Our experiments focus on analyzing uncertainty estimates for selective
prediction rather than maximizing retrieval performance.
\rev{Our evaluation is confined to two architectures
on one benchmark, and these empirical conclusions should be revalidated as further standardized
retrieval benchmarks and model families become available.}
On the methodological side, uncertainty estimation methods
that account for the extreme sparsity of molecular fingerprints,
or that assess epistemic uncertainty directly in the input
space of spectra rather than in learned representations shaped
by the training objective, may yield more informative
estimates than the standard decompositions evaluated here.
\rev{False discovery rate control, for example via conformalized Benjamini--Hochberg
\cite{JMLR_selection_by_prediction}, targets the expected false-discovery proportion in a selected
finite batch.}
Conformal prediction methods that output calibrated prediction
sets \cite{shafer2008tutorial} provide an alternative
paradigm entirely.
\rev{A further limitation is that the evaluated MassSpecGym candidate sets include the reference
structure by construction. Selective prediction when the true molecule is absent from the candidate
pool remains an important open-world setting.}

\backmatter

\bmhead{List of abbreviations}

\begin{description}
  \item[AURC] Area Under the Risk--Coverage curve
  \item[$D_{\mathrm{KL}}$] Kullback--Leibler divergence
  \item[FDR] False Discovery Rate
  \item[GeLU] Gaussian Error Linear Unit
  \item[Hit@$K$] Hit rate at rank cutoff $K$
  \item[i.i.d.] independent and identically distributed
  \item[KNN] $k$-Nearest Neighbors
  \item[MC Dropout] Monte Carlo Dropout
  \item[MCES] Maximum Common Edge Subgraph
  \item[MLP] Multi-Layer Perceptron
  \item[MS/MS] tandem Mass Spectrometry
  \item[$m/z$] mass-to-charge ratio
  \item[relAURC] relative Area Under the Risk--Coverage curve
  \item[SGR] Selection with Guaranteed Risk
\end{description}

\section*{Declarations}

\subsection*{Availability of data and materials}
The MassSpecGym benchmark dataset is publicly available
at \url{https://github.com/pluskal-lab/MassSpecGym}~\cite{bushuiev2024massspecgym}.
All code for model training, uncertainty estimation, evaluation,
and figure generation is available at
\url{https://github.com/mkjuergens/Selective-MSMS}. Model checkpoints, predictions, and 
 evaluation results are archived at \url{https://doi.org/10.5281/zenodo.19108280}.
\subsection*{Competing interests}
The authors declare that they have no competing interests.
\subsection*{Funding}
M.J. and W.W. received funding from the Flemish Government under the “Onderzoeksprogramma Artificiële Intelligentie (AI) Vlaanderen” programme.
\subsection*{Authors' contributions}
M.J. conceived and designed the study, conducted all experiments and analyses, and wrote the manuscript.
G.D.W. provided the base model implementations and training code, contributed to methodological discussions,
provided domain expertise and reviewed the manuscript.
M.R. contributed to methodological discussions and reviewed the manuscript.
W.W. supervised the project, acquired funding, contributed to methodological discussions
and reviewed the manuscript.
\clearpage
\onecolumn
\begin{appendices}

\section{Uncertainty metrics}
\label{appendix: uncertainty metrics}

This appendix provides the formal definitions of the second-order and
distance-based scoring functions summarized in
Section~\ref{sec:scoring-functions}.
Throughout, we assume access to $S$ samples
$\boldsymbol{\theta}^{(1)}, \ldots, \boldsymbol{\theta}^{(S)}
\sim q(\boldsymbol{\theta}\mid\mathbf{x})$,
obtained via a Bayesian approximation of $q(\boldsymbol{\theta}\mid\mathbf{x})$.
All expectations and variances with respect to $q(\boldsymbol{\theta}\mid\mathbf{x})$ are approximated by their
sample counterparts.

\subsection{Bitwise uncertainty decomposition}
\label{appendix:bitwise}

\textbf{Information-theoretic decomposition.} Assuming independence between fingerprint bits, we model each bit probability via the
marginal distribution $\theta_d \sim q_d(\theta_d\mid\mathbf{x})$,
hence for the fingerprint bits $y_d\mid\theta_d \sim \mathrm{Ber}(\theta_d)$.
Let $\overline{\theta}_d = \mathbb{E}_{q}[\theta_d]$ denote the posterior
mean for bit~$d$. Applying the information-theoretic
decomposition~\cite{depeweg2018decomposition} to the mean predicted bit probability,
we can derive expressions for total, aleatoric and epistemic uncertainty:
\begin{equation*}
  \underbrace{\mathbb{H}(\overline{\theta}_d)}_{u_{\mathrm{tot}}^{(d)}}
  = \underbrace{\mathbb{E}_{q}\!\bigl[\mathbb{H}(y_d \mid \theta_d)\bigr]}_{u_{\mathrm{al}}^{(d)}}
  + \underbrace{\mathbb{E}_{q}\!\bigl[D_{\mathrm{KL}}(\theta_d \,\|\, \overline{\theta}_d)\bigr]}_{u_{\mathrm{ep}}^{(d)}}\,,
  \qquad d = 1, \ldots, D\,,
\end{equation*}
where $\mathbb{H}(p) = -p\log p - (1-p)\log(1-p)$ is the binary entropy.
Following~\cite{sale2024labelwisealeatoricepistemicuncertainty}, we aggregate the bitwise uncertainties
to fingerprint-level uncertainties by summing over all bits:
\begin{equation*}
  \kappa_{\mathrm{bit}}^{(\cdot)}(\mathbf{x})
  = -\sum_{d=1}^{D} u_{(\cdot)}^{(d)}\,,
\end{equation*}
where $(\cdot) \in \{\mathrm{tot}, \mathrm{al}, \mathrm{ep}\}$ and the
negation ensures the convention that higher values indicate greater
confidence.\\

\subsection{Retrieval-level uncertainty decomposition}
\label{appendix:retrieval}
\bmhead{Information-theoretic decomposition} For the uncertainty on the candidate level, we 
look at the distribution over candidate probabilities.
Each sample $\boldsymbol{\theta}^{(r)}$ induces a candidate distribution
$p_j^{(r)}(\mathbf{x})$ via
Eq.~(\ref{eq:candidate_prob}).
Let
$\overline{p}_j(\mathbf{x}) = \frac{1}{S}\sum_{r=1}^{S}
p_j^{(r)}(\mathbf{x})$
denote the expected candidate probability. Following the same information-theoretic decomposition as 
for the fingerprint-level uncertainty, but applied to the candidate probabilities,
 the empirical decomposition for the retrieval-level uncertainty is given as
\begin{align*}
  \kappa_{\mathrm{ret}}^{\mathrm{tot}}(\mathbf{x})
    &= \sum_{j=1}^{M_i} \overline{p}_j(\mathbf{x})\,
       \log \overline{p}_j(\mathbf{x})\, \\
  \kappa_{\mathrm{ret}}^{\mathrm{al}}(\mathbf{x})
    &= \frac{1}{S}\sum_{r=1}^{S} \sum_{j=1}^{M_i}
       p_j^{(r)}(\mathbf{x})\,
       \log p_j^{(r)}(\mathbf{x}),\\
  \kappa_{\mathrm{ret}}^{\mathrm{ep}}(\mathbf{x})
    &= \kappa_{\mathrm{ret}}^{\mathrm{tot}}(\mathbf{x})
     - \kappa_{\mathrm{ret}}^{\mathrm{al}}(\mathbf{x}),
\end{align*}
where the sign convention ensures that higher values correspond to lower
uncertainty.
Note that $\kappa_{\mathrm{ret}}^{\mathrm{tot}}$ equals the negative
entropy of the expected candidate distribution, while
$\kappa_{\mathrm{ret}}^{\mathrm{al}}$ equals the negative expected
conditional entropy, so that $-\kappa_{\mathrm{ret}}^{\mathrm{ep}}$
corresponds to the mutual information between $\boldsymbol{\theta}$ and the
predicted candidate.

\bmhead{Rank variance}
For each spectrum $\mathbf{x}$, let
$\mathcal{S}_K(\mathbf{x})$ denote the top-$K$ candidate set under the
mean prediction.
\begin{revblock}
For each candidate
$\mathbf{c}_j \in \mathcal{S}_K(\mathbf{x})$, let
$\rho_j^{(r)}$ denote its rank when candidates are re-ranked using
sample $\boldsymbol{\theta}^{(r)}$.
The rank-variance scoring function is defined as
\begin{equation*}
  \kappa_{\mathrm{rank}}(\mathbf{x})
  = -\frac{1}{K} \sum_{\mathbf{c}_j \in \mathcal{S}_K(\mathbf{x})}
    \mathrm{Var}_{r}\!\bigl[\rho_j^{(r)}\bigr]\,,
\end{equation*}
\end{revblock}
where the variance is taken over the $S$ posterior samples.
High rank variance indicates that the relative ordering of top candidates
is sensitive to parameter uncertainty, suggesting unreliable retrieval.

\subsection{Distance-based scoring functions}
\label{appendix:distance}

Let $\mathbf{h}(\mathbf{x}) \in \mathbb{R}^{d_h}$ denote the
penultimate-layer representation of the model for input $\mathbf{x}$,
and let
$\mathcal{H}_{\mathrm{train}} = \{\mathbf{h}(\mathbf{x}_i)\}_{i=1}^{N}$
be the set of training representations.\\

\bmhead{KNN distance}
Let $\mathbf{h}_1^{\mathrm{nn}}, \ldots, \mathbf{h}_k^{\mathrm{nn}}$ be
the $k$ nearest elements of $\mathcal{H}_{\mathrm{train}}$ to
$\mathbf{h}(\mathbf{x})$ in Euclidean distance. Then
\begin{equation*}\label{eq:knn}
  \kappa_{\mathrm{knn}}(\mathbf{x})
  = -\frac{1}{k}\sum_{i=1}^{k}
    \bigl\|\mathbf{h}(\mathbf{x}) - \mathbf{h}_i^{\mathrm{nn}}\bigr\|_2\,.
\end{equation*}

\bmhead{Mahalanobis distance}
Let $\boldsymbol{\mu} = \frac{1}{N}\sum_{i=1}^{N}\mathbf{h}(\mathbf{x}_i)$
and
$\boldsymbol{\Sigma} = \frac{1}{N}\sum_{i=1}^{N}
(\mathbf{h}(\mathbf{x}_i) - \boldsymbol{\mu})
(\mathbf{h}(\mathbf{x}_i) - \boldsymbol{\mu})^{\!\top}$
be the sample mean and covariance of the training representations. Then
\begin{equation*}\label{eq:mahalanobis}
  \kappa_{\mathrm{mah}}(\mathbf{x})
  = -\sqrt{(\mathbf{h}(\mathbf{x}) - \boldsymbol{\mu})^{\!\top}
    \boldsymbol{\Sigma}^{-1}
    (\mathbf{h}(\mathbf{x}) - \boldsymbol{\mu})}\,.
\end{equation*}

\bmhead{Configuration and hyperparameter search}
We evaluate both scoring functions under the Deep Ensemble
trained with ranking loss and with focal loss.
We use the
$1024$-dimensional encoder representation 
with optional PCA dimensionality reduction to 64 or 128
components, and neighbourhood sizes
$k \in \{1, 10, 50, 100\}$ for
$\kappa_{\mathrm{knn}}$.
All embeddings are $\ell_2$-normalised.
The best relAURC across all configurations is $0.886$
(k-NN, encoder, PCA to 128 dimensions, $k{=}100$,
Hit@20, focal loss), compared to $0.30$ for confidence.

\section{Risk control algorithm}
\label{appendix:risk-control}

This appendix provides the algorithmic details of the risk control
procedure described in Section~\ref{sec:risk-control}, namely, the global risk control method SGR \cite{geifman2017selective}.
To this end, we assume access to a 
calibration set $\mathcal{D}_{\mathrm{cal}} = \{(\mathbf{x}_i, \mathbf{y}_i,
\mathcal{C}_i)\}_{i=1}^{n}$ of $n$ i.i.d.\ samples from the data distribution~$P$.
In the following, let $\kappa: \mathcal{X} \rightarrow \mathbb{R}$ be a scoring function.
We start by sorting the calibration samples by ascending confidence:
$\kappa(\mathbf{x}_{(1)}) \leq \cdots \leq \kappa(\mathbf{x}_{(n)})$.
For any candidate threshold $\tau$, one then defines the accepted subset
$\mathcal{A}(\tau) = \{i : \kappa(\mathbf{x}_i) \geq \tau\}$ with
$n_\tau = |\mathcal{A}(\tau)|$, and the empirical selective risk on this subset:
\begin{equation*}
  \hat{r}(\tau) =
  \frac{\sum_{i \in \mathcal{A}(\tau)}
    \ell_K(\mathbf{x}_i, \mathbf{y}_i)}{n_\tau}\,.
\end{equation*}
Since $\ell_K \in \{0, 1\}$, the number of errors
$k_\tau = \sum_{i \in \mathcal{A}(\tau)} \ell_K(\mathbf{x}_i,
\mathbf{y}_i)$ among $n_\tau$ accepted samples is a sufficient statistic.
The Clopper--Pearson upper bound~\cite{gascuel1992distribution}
$B^*(\hat{r}, \delta', n_\tau)$ is the solution~$b$ of
\begin{equation*}
  \sum_{j=0}^{k_\tau}
  \binom{n_\tau}{j}\, b^{j}\,(1 - b)^{n_\tau - j} = \delta'\,,
\end{equation*}
which yields the tightest $b$ such that
$\mathbb{P}\bigl(R(f, g_\tau) > b\bigr) < \delta'$ for a fixed
threshold~$\tau$.
The SGR algorithm performs a binary search over threshold indices
$z \in \{1, \ldots, n\}$, examining at most
$k = \lceil \log_2 n \rceil$ thresholds.
At iteration~$i$, it sets $\tau = \kappa(\mathbf{x}_{(z)})$,
computes the empirical risk $\hat{r}(\tau)$, and evaluates the bound
$b^*_i = B^*(\hat{r}, \delta/k, n_\tau)$ with adjusted significance
$\delta/k$.
If $b^*_i < r^*$, the search moves toward higher coverage (lower $z$),
otherwise toward lower coverage (higher $z$).
The output is the final threshold~$\tau^*$ and bound~$b^*$, satisfying
the guarantee 
\begin{equation*}
\mathbb{P}\bigl(R(f, g_{\tau^*}) > b^*\bigr) < \delta.
\end{equation*}


\section{Experimental details}
\label{sec: experimental details}

\subsection{Dataset characterisation}
\label{sec: dataset characterisation}

\begin{figure}[t]
\includegraphics[width=\textwidth]{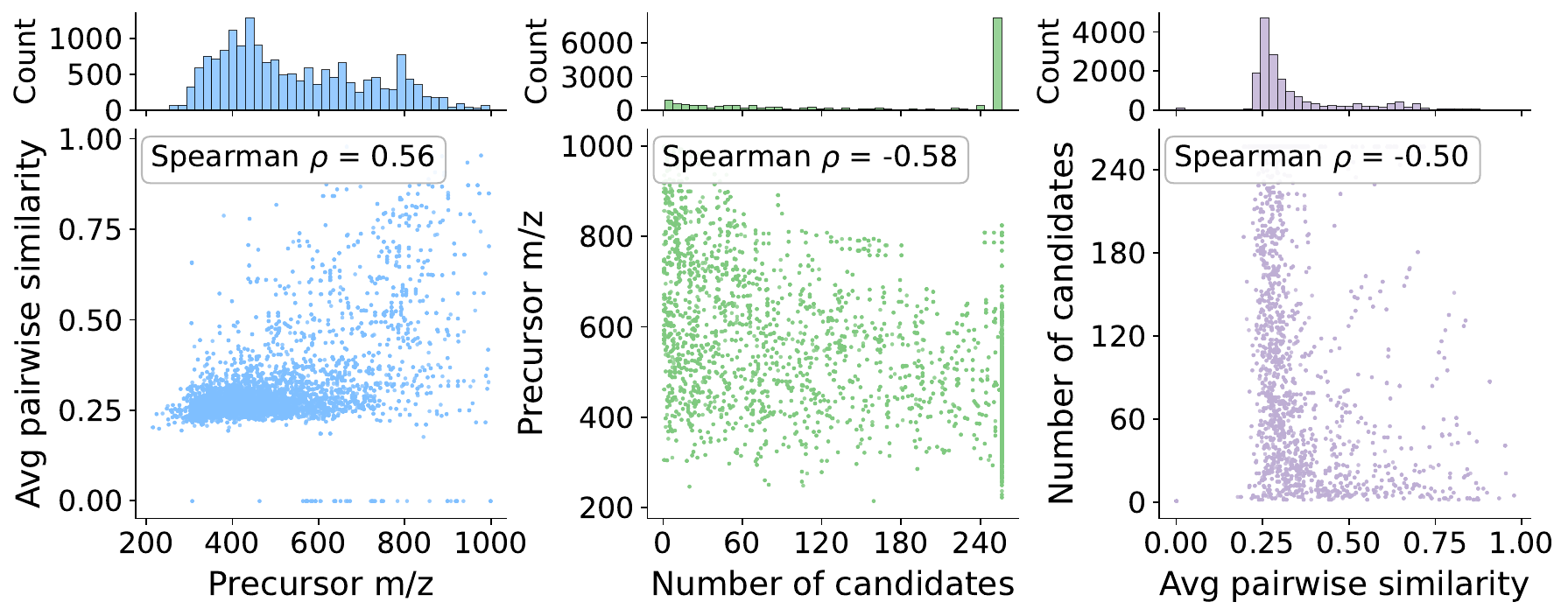}
\caption{Pairwise scatter plots of precursor mass, average pairwise cosine
similarity and number of candidates per spectrum in the candidate sets of the
MassSpecGym test set. Spearman rank correlations are shown in each panel.
Histograms on the top and right axes show the marginal distribution of each
variable. Larger candidate sets arise for lower precursor masses and contain
more structurally diverse candidates.}
\label{fig:candidate-stats}
\end{figure}

The MassSpecGym test set contains $17\,556$ spectra with candidate set
sizes ranging from $1$ to the benchmark cap of $256$.
The recently reported retrieval leakage on MassSpecGym stems from inconsistent canonical and
non-canonical SMILES representations in candidate lists \cite{liu2026massspecgym}. As here we rank Morgan fingerprints,  this specific leakage mechanism cannot
be exploited in this setup.
Figure~\ref{fig:candidate-stats} characterises the candidate sets obtained by filtering
by molecular formula 
through three pairwise relationships.
Candidate set size correlates negatively with precursor mass:
heavier molecules have fewer database matches, yielding smaller
candidate sets.
At the same time, larger candidate sets exhibit lower average pairwise
cosine similarity among their members, indicating greater structural
diversity.
These two effects are linked: low-mass molecules are more densely
represented in chemical databases, producing large candidate sets
of structurally diverse molecules, while high-mass molecules yield
small, structurally homogeneous candidate sets.
This variation has direct consequences for both retrieval difficulty
and uncertainty estimation, as analyzed in
Section~\ref{sec:candidate-set-analysis}.

\subsection{Model architecture and training}
\label{sec: model-architecture}

The base model is a three-layer MLP following the fingerprint predictor
of De~Waele et~al.~\cite{de2026small}
The input spectrum is binned with binwidth $0.1\,\text{Da}$, yielding a vector
$\mathbf{x} \in \mathbb{R}^{10\,050}$.
Three fully connected layers with hidden dimension~$1024$, GeLU
activations, layer normalisation and dropout ($p = 0.3$) map the input
to a $1024$-dimensional representation.
A linear output layer
$W_{\mathrm{out}} \in \mathbb{R}^{4096 \times 1024}$ followed by a
component-wise sigmoid produces the fingerprint prediction
$\boldsymbol{\theta} = \sigma(W_{\mathrm{out}} f(\mathbf{x}))
\in [0,1]^{4096}$.
We train two model variants.
The first is trained with focal loss~\cite{lin2017focal} on
individual bits ($\gamma = 2.0$, learning rate $3 \times 10^{-4}$).
The second is trained with the contrastive ranking loss
(Eqn.~\ref{eq:ranking_loss}) using cosine similarity and temperature
$T = 0.003$ (learning rate $1 \times 10^{-4}$).
Both variants use the Adam optimizer, a batch size of~$64$,
gradient clipping at~$1.0$ and are trained
for a maximum of~$50$ epochs.
Model selection is based on the best validation Hit@20 under cosine
similarity retrieval.

\subsection{Posterior approximation methods}
\label{appendix:posterior-methods}

\bmhead{Deep Ensemble}
Five copies of the base model are trained independently with different
random initialisations (seeds~$42$--$46$) and data-loader shuffles.
Model selection is performed per member independently.

\bmhead{MC~Dropout}
A single trained model (one ensemble member) is evaluated with dropout
enabled at test time, using $S = 50$ stochastic forward passes with
the training dropout rate ($p = 0.3$).

\bmhead{Laplace approximation}
A diagonal last-layer Laplace approximation~\cite{daxberger2021laplace}
is fitted to a single trained model.
The posterior is a Gaussian over the final linear layer
$W_{\mathrm{out}}$, with precision estimated via the diagonal
generalised Gauss--Newton (GGN) approximation accumulated over
$200$~training batches.
The prior precision is tuned via marginal likelihood maximisation
over a logarithmically spaced grid.
At test time, $S = 50$ weight samples are drawn and
passed through the sigmoid to produce fingerprint probability samples.

{\color{blue}\section{Supplementary experimental results}}
\label{sec:supplementary-results}
\label{sec: further experimental results}
\label{appendix:additional-experiments}
{\color{blue}\subsection{Fingerprint-level selective prediction}}
\label{sec: similarity loss}
The main analysis in Section~\ref{sec: risk-coverage analysis} evaluates
selective prediction with the retrieval loss
$\ell_K = 1 - \mathrm{Hit}@K$, where retrieval-level scoring functions
such as confidence and aleatoric uncertainty achieve the best
risk-coverage tradeoffs. Here we evaluate with fingerprint-level
similarity losses to test whether the level-alignment principle extends
bidirectionally.

\bmhead{Similarity losses}
\begin{revblock}
Let $\overline{\boldsymbol{\theta}} \in [0,1]^D$ denote the ensemble-mean
predicted probabilities and
$\hat{\mathbf{y}} = \mathbb{I}[\overline{\boldsymbol{\theta}} > 0.5]
\in \{0,1\}^D$ the binarised prediction,
with ground truth $\mathbf{y} \in \{0,1\}^D$.
\end{revblock}
We consider three similarity measures, each of which can be evaluated
with either the continuous or the discrete predicted fingerprint:
\begin{align*}
    \mathrm{Tanimoto}(\mathbf{a}, \mathbf{y})
    &= \frac{\sum_{j} a_j \, y_j}
            {\sum_{j} a_j + \sum_{j} y_j
             - \sum_{j} a_j \, y_j}\,,
\\[4pt]
  \mathrm{Cosine}(\mathbf{a}, \mathbf{y})
    &= \frac{\mathbf{a}^{\top} \mathbf{y}}
            {\|\mathbf{a}\|_2 \, \|\mathbf{y}\|_2}\,,
 \\[4pt]
  \mathrm{Hamming}(\mathbf{a}, \mathbf{y})
    &= 1 - \frac{1}{D}\sum_{j=1}^{D}
      \mathbb{I}[a_j \neq y_j]\,,
\end{align*}
where \rev{$\mathbf{a} = \overline{\boldsymbol{\theta}}$} for the continuous prediction and
$\mathbf{a} = \hat{\mathbf{y}}$ for the discrete prediction.
The corresponding losses are
$\ell_{\mathrm{sim}} = 1 - \mathrm{sim}(\mathbf{a}, \mathbf{y})$.
Hamming loss requires binary inputs and is therefore always evaluated
on~$\hat{\mathbf{y}}$.

\bmhead{Continuous evaluation}
When Tanimoto and cosine similarity are evaluated on the continuous
probabilities, the loss is sensitive not only to which bits cross the
binarisation threshold, but also to how far each prediction is from it.
Table~\ref{tab:aurc-fingerprint-continuous} (left) shows that bitwise
total uncertainty achieves the lowest relAURC for both Tanimoto and
cosine loss, reversing the retrieval-level dominance observed for hit
rate. This pattern is consistent across all three posterior
approximation methods.
The explanation is that continuous Tanimoto and bitwise total uncertainty
measure the same property: both are determined by how peaked the
per-bit probabilities are near zero or one, rather than by which side
of the decision boundary they fall on.
\bmhead{Discrete evaluation}
To test whether this alignment reflects genuine substructure-level
prediction quality or merely probability calibration, we repeat
the analysis with the binarised fingerprint
(Table~\ref{tab:aurc-fingerprint-continuous}, right).
The advantage of bitwise scoring functions largely vanishes:
bitwise entropy, which is entirely about distance-from-boundary, loses
its predictive power.
This reveals that the apparent level alignment in the continuous setting
was driven by a coupling between bitwise entropy and probability
calibration, rather than by genuine predictive value at the
substructure level.

\bmhead{Hamming loss}
Nearly all scoring functions perform worse than random rejection
(relAURC~$> 1$), because Hamming weights all bits equally while
both bitwise and retrieval-level scoring functions concentrate
their signal on the sparse minority of active bits.
The sole exception is bitwise epistemic uncertainty, which captures
disagreement on inactive bits that other scores ignore:
on these bits the aleatoric component vanishes, so any inter-sample disagreement manifests
entirely as epistemic uncertainty.
This gives it a more uniform sensitivity across all bits, matching
the uniform weighting of Hamming loss.
\begin{table}[t]
\caption{%
  Selective prediction quality for fingerprint-level similarity losses,
  reported as relAURC (Eqn.~\ref{eq:rel_aurc}). Lower is better.
  Left: continuous evaluation on predicted
  \rev{probabilities~$\overline{\boldsymbol{\theta}}$.}
  Right: discrete evaluation on binarised
  \rev{fingerprints~$\hat{\mathbf{y}} = \mathbb{I}[\overline{\boldsymbol{\theta}} > 0.5]$.}
  Hamming always uses discrete inputs.
  Results are shown for the Deep Ensemble ($S=5$) trained with focal loss,
  and the experimental setup as described in Section \ref{sec:experimental-setup}.}
\label{tab:aurc-fingerprint-continuous}
\centering
\footnotesize
\setlength{\tabcolsep}{4pt}
\renewcommand{\arraystretch}{1.05}
\begin{tabular*}{\textwidth}{@{\extracolsep{\fill}}l rr c rr r@{}}
\toprule
 & \multicolumn{2}{c}{Continuous}
 & \phantom{a}
 & \multicolumn{2}{c}{Discrete}
 & \\
\cmidrule(lr){2-3}\cmidrule(lr){5-6}
$\kappa$
 & Tan. & Cos.
 &
 & Tan. & Cos.
 & Ham. \\
\midrule
\multicolumn{7}{@{}l}{\textit{Fingerprint-level}} \\
\quad $\kappa_{\mathrm{bit}}^{\mathrm{tot}}$\; total
  & \textbf{0.367} & \textbf{0.712}
  && 0.818 & 0.974
  & 1.335 \\
\quad $\kappa_{\mathrm{bit}}^{\mathrm{al}}$\; aleatoric
  & 0.534 & 0.713
  && 0.853 & 0.971
  & 1.171 \\
\quad $\kappa_{\mathrm{bit}}^{\mathrm{ep}}$\; epistemic
  & 1.142 & 1.150
  && 1.099 & 1.061
  & \textbf{0.777} \\
\midrule
\multicolumn{7}{@{}l}{\textit{Retrieval-level}} \\
\quad $\kappa_{\mathrm{ret}}^{\mathrm{al}}$\; aleatoric
  & 0.755 & 0.768
  && \textbf{0.809} & \textbf{0.867}
  & 1.156 \\
\quad $\kappa_{\mathrm{conf}}$\; confidence
  & 0.826 & 0.835
  && 0.859 & 0.908
  & 1.130 \\
\quad $\kappa_{\mathrm{rank}}$\; $K\!=\!5$
  & 0.852 & 0.871
  && 0.849 & 0.900
  & 1.062 \\
\midrule
\multicolumn{7}{@{}l}{\textit{Other}} \\
\quad $|\mathcal{C}|$\; num.\ candidates
  & 0.921 & 0.948
  && 1.002 & 1.033
  & 1.255 \\
\bottomrule
\end{tabular*}
\end{table}

\subsection{Scoring function correlations}
\label{sec:correlation-analysis}

\begin{figure}[t]
  \centering
  \includegraphics[width=.75\textwidth]{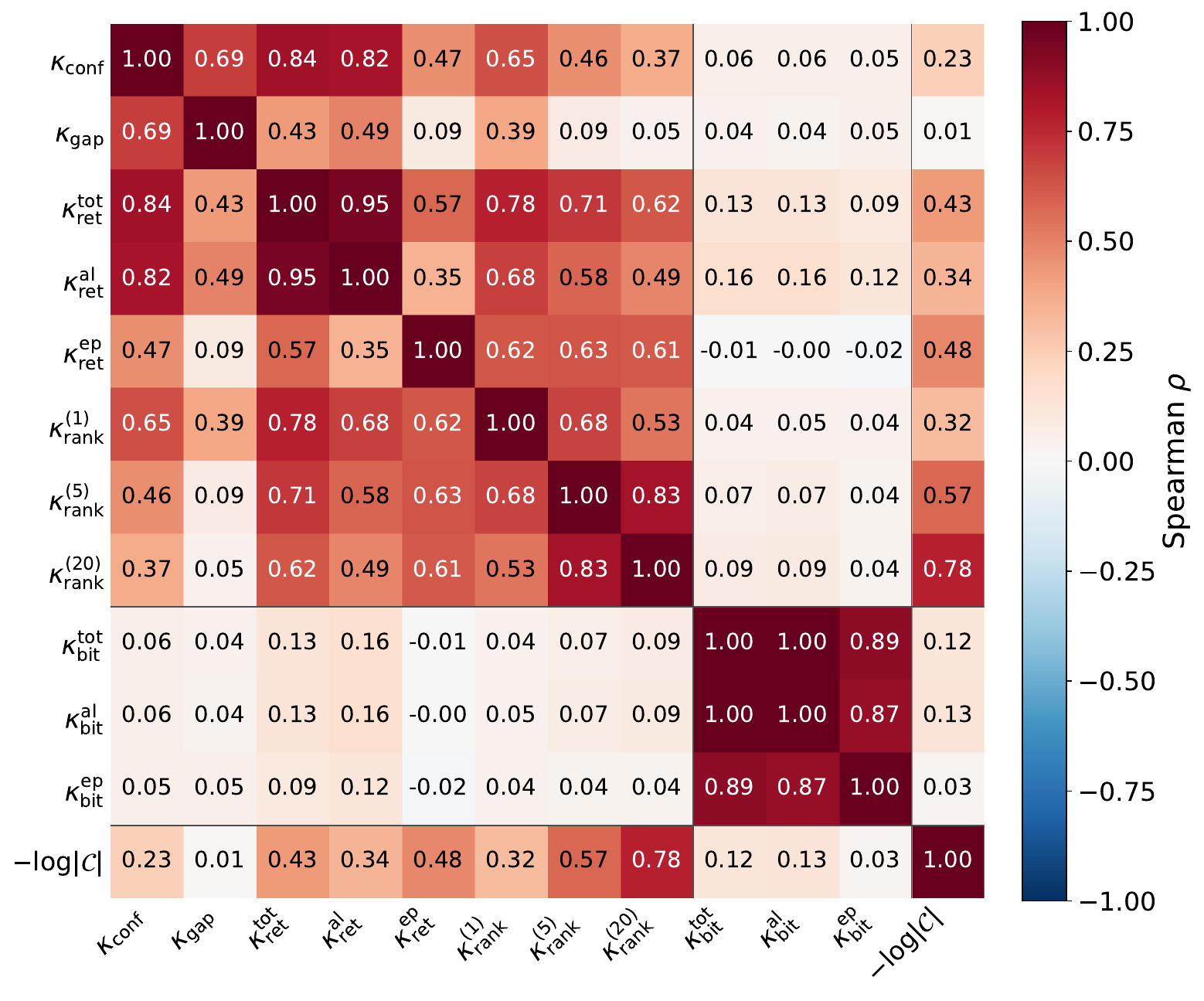}
  \caption{\rev{Pairwise Spearman rank correlations among the main scoring
  functions for the MLP Deep Ensemble trained with ranking loss, evaluated
  on the official formula-filtered MassSpecGym test candidate sets at
  $T_{\mathrm{eval}}=0.003$. All quantities are oriented so that higher
  values indicate greater confidence; the candidate-set-size criterion is
  therefore shown as $-\log |\mathcal{C}|$. Horizontal and vertical lines
  separate retrieval-level scores, fingerprint-level scores, and the
  candidate-set-size criterion.}}
  \label{fig:correlation-heatmaps}
\end{figure}

\begin{revblock}
Figure~\ref{fig:correlation-heatmaps} shows pairwise Spearman rank
correlations among the main scoring functions for the MLP Deep
Ensemble trained with ranking loss. Results are computed on the official
formula-filtered MassSpecGym test candidate sets at
$T_{\mathrm{eval}}=0.003$. All quantities are oriented such that higher
values indicate greater confidence: uncertainty measures are negated, and
candidate-set size is represented by $-\log|\mathcal{C}|$.

The retrieval-level scores are correlated but not interchangeable. Total and
aleatoric retrieval uncertainty are strongly correlated
($\rho=0.95$), and both are closely associated with confidence
($\rho=0.84$ and $\rho=0.82$, respectively). The score
gap also correlates with confidence ($\rho=0.69$), but is only weakly
associated with retrieval epistemic uncertainty ($\rho=0.09$). The relationship of rank
variance to the other scores changes with the retrieval cutoff: its
correlation with total uncertainty decreases from $\rho=0.78$ for $K=1$
to $\rho=0.62$ for $K=20$, whereas its correlation with
$-\log|\mathcal{C}|$ increases from $\rho=0.32$ to $\rho=0.78$.
Thus, rank variance is increasingly associated with candidate-set size at
broader cutoffs. The stratified analysis in
Section~\ref{sec:candidate-set-analysis}, however, shows that rank variance
remains informative when candidate-set size is fixed and is therefore not
solely a proxy for set size.

The fingerprint-level scores are strongly correlated with one another but
only weakly with the retrieval-level scores ($|\rho|\leq 0.16$). This
supports the conclusion that uncertainty is most informative when evaluated
at the level of the target loss.
\end{revblock}

\clearpage
\begin{revblock}
\subsection{Transformer spectrum encoder}
\label{appendix:transformer}
\begin{figure}[htbp]
  \color{blue}
  \centering
  \includegraphics[width=\textwidth]{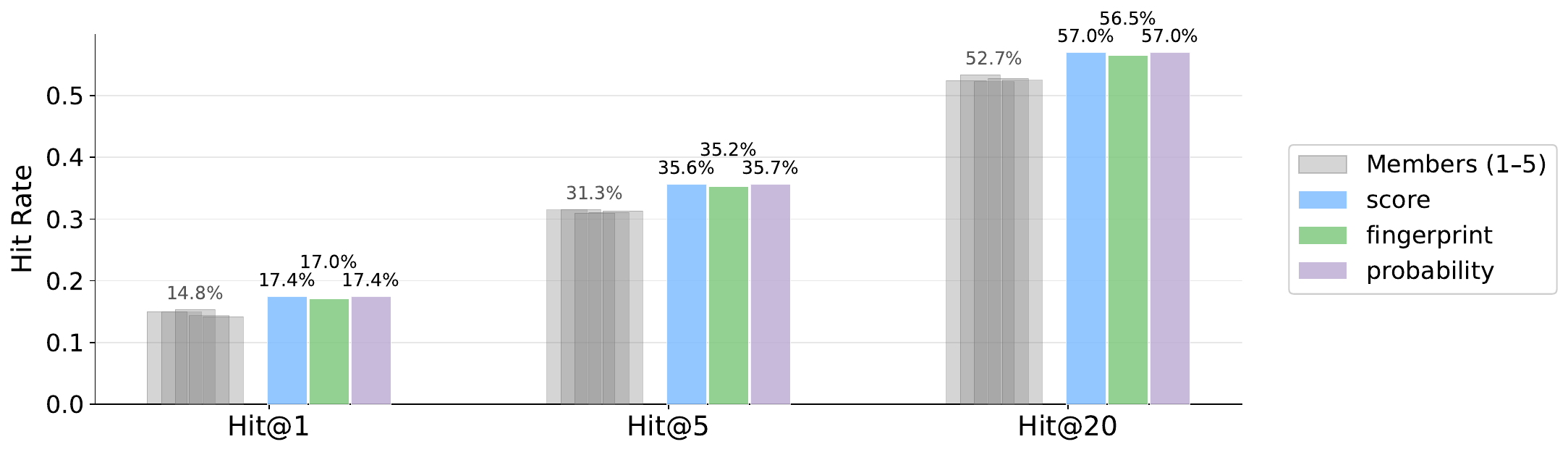}
  \caption{\rev{Average $\mathrm{Hit}@K$ rate for $K\in\{1,5,20\}$ for the individual transformer ensemble
  members (grey) and their aggregate (blue, green, purple), for a Deep Ensemble of $S=5$ members on the
  MassSpecGym test set with candidates filtered by molecular formula.}}
  \label{fig:transformer-aggregation}
\end{figure}
To test whether our conclusions depend on the MLP fingerprint predictor, we repeat the main
selective-prediction analysis with a transformer spectrum encoder, following the 
architecture of~\citet{de2026small}. In contrast to the binned-spectrum representation, the transformer
represents each spectrum by its most intense peaks, where each peak is tokenised through
sinusoidal encodings of its $m/z$ and intensity.
The model is trained with the same contrastive ranking loss
 and all remaining hyperparameters and the training procedure as for the MLP setup. We again
form a Deep Ensemble of $S=5$ independently trained members. As for the MLP, aggregating over the
ensemble members improves the hit rate over the individual members, as can be seen in Figure~\ref{fig:transformer-aggregation}.

\bmhead{Risk-coverage} 
Table~\ref{tab:transformer-relaurc} reports the relAURC values for the
transformer ensemble, and Figure~\ref{fig:transformer-rc} shows the
corresponding risk--coverage curves. The transformer again showsthat selection criteria should 
be task-aligned: retrieval-level scores clearly outperform
fingerprint-level uncertainty, whose relAURC remains close to random
rejection.
The strongest criterion depends on the retrieval cutoff. Confidence performs
best for $K=1$, followed by total retrieval
uncertainty. Total retrieval uncertainty performs best for $K=5$ and $K=20$
 and consistently outperforms its
aleatoric and epistemic components. Rank variance remains competitive for larger $K$: for $K=5$, it is the second-best criterion,
 but, unlike for the MLP, it does not outperform total
uncertainty. Thus, the transformer confirms the overall advantage of
retrieval-level total uncertainty while showing that the relative ordering of
confidence and rank stability depends on the architecture and cutoff.
\begin{table}[htbp]
  \color{blue}
  \centering
  \caption{\rev{relAURC (lower is better) of all scoring functions for the transformer Deep Ensemble
  ($S=5$) trained with ranking loss, on the MassSpecGym test set with candidates filtered by molecular
  formula.}}
  \label{tab:transformer-relaurc}
  \footnotesize
  \setlength{\tabcolsep}{4pt}
  \renewcommand{\arraystretch}{1.05}
  \begin{tabular*}{\textwidth}{@{\extracolsep{\fill}}l rrr@{}}
  \toprule
  $\kappa$ & $\ell_1$ & $\ell_5$ & $\ell_{20}$ \\
  \midrule
  \multicolumn{4}{@{}l}{\textit{Retrieval-level}} \\
  \quad $\kappa_{\mathrm{conf}}$\; confidence          & \textbf{0.356} & 0.367 & 0.420 \\
  \quad $\kappa_{\mathrm{gap}}$\; score gap            & 0.384 & 0.513 & 0.606 \\
  \quad $\kappa_{\mathrm{ret}}^{\mathrm{tot}}$\; total      & 0.365 & \textbf{0.317} & \textbf{0.320} \\
  \quad $\kappa_{\mathrm{ret}}^{\mathrm{al}}$\; aleatoric   & 0.377 & 0.342 & 0.369 \\
  \quad $\kappa_{\mathrm{ret}}^{\mathrm{ep}}$\; epistemic   & 0.501 & 0.542 & 0.591 \\
  \quad $\kappa_{\mathrm{rank}}$\; $K\!=\!1$           & 0.446 & 0.404 & 0.401 \\
  \quad $\kappa_{\mathrm{rank}}$\; $K\!=\!5$           & 0.579 & 0.429 & 0.347 \\
  \quad $\kappa_{\mathrm{rank}}$\; $K\!=\!20$          & 0.609 & 0.500 & 0.409 \\
  \midrule
  \multicolumn{4}{@{}l}{\textit{Fingerprint-level}} \\
  \quad $\kappa_{\mathrm{bit}}^{\mathrm{tot}}$\; total      & 0.933 & 0.954 & 0.957 \\
  \quad $\kappa_{\mathrm{bit}}^{\mathrm{al}}$\; aleatoric   & 0.938 & 0.965 & 0.971 \\
  \quad $\kappa_{\mathrm{bit}}^{\mathrm{ep}}$\; epistemic   & 0.921 & 0.910 & 0.913 \\
  \midrule
  \multicolumn{4}{@{}l}{\textit{Other}} \\
  \quad $|\mathcal{C}|$\; num.\ candidates & 0.634 & 0.549 & 0.484 \\
  \bottomrule
  \end{tabular*}
\end{table}

\begin{figure}[htbp]
  \color{blue}
  \centering
  \includegraphics[width=\textwidth]{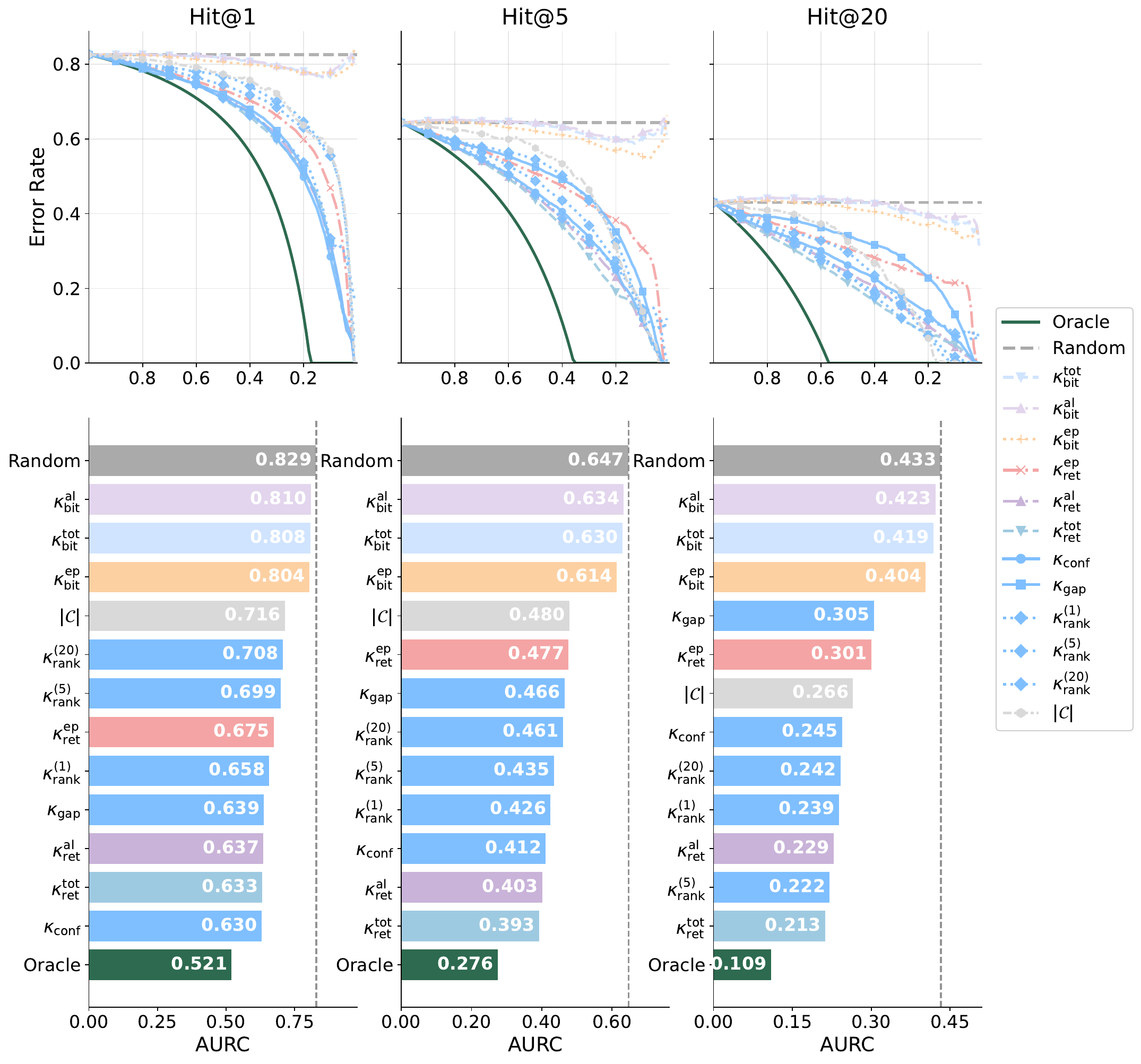}
  \caption{\rev{Risk--coverage analysis for the transformer spectrum encoder ensemble on the MassSpecGym test set,
  for $\ell_K = 1 - \mathrm{Hit@}K$, $K\in\{1,5,20\}$.}}
  \label{fig:transformer-rc}
\end{figure}

\bmhead{Risk-controlled selection}
We repeat the SGR analysis from Section~\ref{sec:risk-control-results} for the
transformer Deep Ensemble. For each retrieval cutoff, the scoring function is
selected independently on the official validation fold by relAURC. This selects
confidence for Hit@1 and retrieval-level total uncertainty for Hit@5 and Hit@20.
The SGR threshold is then calibrated on one half of the test set and evaluated on
the held-out half, using the same split, target risks, and $\delta=0.001$ as for
the MLP.

Figure~\ref{fig:transformer-sgr-coverage} shows the coverage and held-out empirical
risk for the evaluated scoring functions. Table~\ref{tab:transformer-sgr-results}
reports the numerical results for the validation-selected score at each cutoff.
The stronger base retrieval performance of the transformer permits higher coverage
than the MLP, particularly for Hit@20. At a target risk of $r^*=0.5$, the selected
scores retain 18.25\%, 55.80\%, and 99.97\% of the evaluation spectra for
Hit@1, Hit@5, and Hit@20, respectively. At every feasible operating point reported
in the table, the held-out empirical risk remains below the target risk. 

\begin{figure}[htbp]
  \color{blue}
  \centering
  \includegraphics[width=\textwidth]{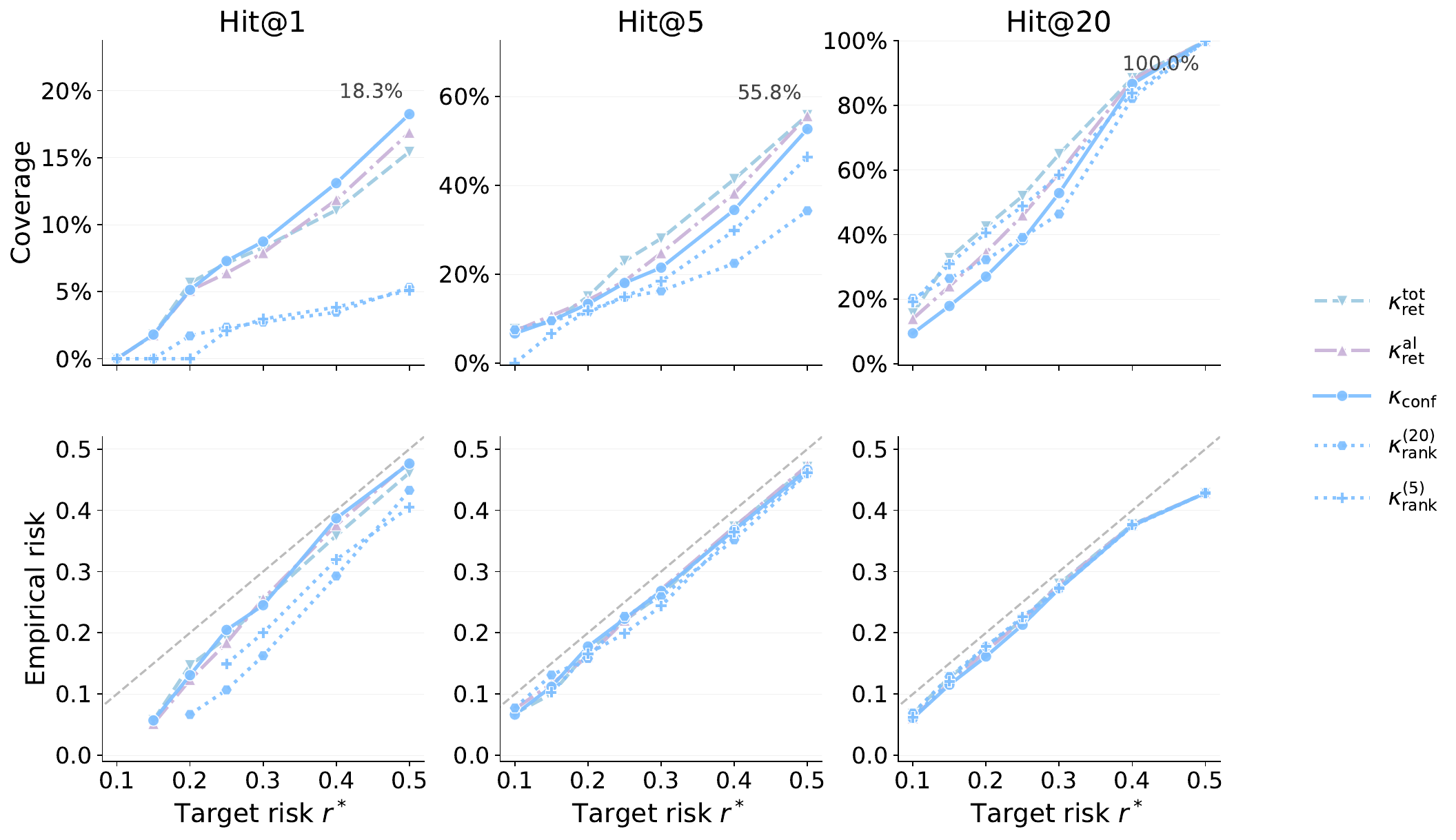}
  \caption{\rev{Risk-controlled annotation for the transformer Deep Ensemble with
  $\delta=0.001$ and target risks $r^*$. Thresholds are calibrated on one half of
  the MassSpecGym test set and evaluated on the held-out half, using formula-filtered
  candidates. \textbf{Top}: coverage at each target risk for
  $K\in\{1,5,20\}$. \textbf{Bottom}: held-out empirical risk versus target risk.}}
  \label{fig:transformer-sgr-coverage}
\end{figure}

\begin{table}[htbp]
  \color{blue}
  \centering
  \caption{\rev{Numerical SGR results for the transformer Deep Ensemble. For each
  retrieval cutoff, the scoring function is selected on the official validation
  fold: confidence for Hit@1 and retrieval-level total uncertainty for Hit@5 and
  Hit@20. Coverage and empirical risk are percentages evaluated on the held-out
  half of the test set after threshold calibration on the other half, with
  $\delta=0.001$. A dash indicates that no non-empty feasible selection was found.}}
  \label{tab:transformer-sgr-results}
  \footnotesize
  \setlength{\tabcolsep}{4pt}
  \renewcommand{\arraystretch}{1.05}
  \begin{tabular*}{\textwidth}{@{\extracolsep{\fill}}l rr rr rr@{}}
    \toprule
    & \multicolumn{2}{c}{Hit@1, $\kappa_{\mathrm{conf}}$}
    & \multicolumn{2}{c}{Hit@5, $\kappa_{\mathrm{ret}}^{\mathrm{tot}}$}
    & \multicolumn{2}{c}{Hit@20, $\kappa_{\mathrm{ret}}^{\mathrm{tot}}$} \\
    \cmidrule(lr){2-3}\cmidrule(lr){4-5}\cmidrule(lr){6-7}
    $r^*$ & Coverage & Risk & Coverage & Risk & Coverage & Risk \\
    \midrule
    0.10 & ---   & ---   &  7.67 &  6.84 & 15.54 &  6.74 \\
    0.15 &  1.80 &  5.70 &  9.23 & 10.00 & 32.71 & 12.75 \\
    0.20 &  5.14 & 13.08 & 14.99 & 16.95 & 42.48 & 17.35 \\
    0.25 &  7.29 & 20.47 & 22.98 & 22.01 & 51.91 & 22.16 \\
    0.30 &  8.74 & 24.51 & 28.05 & 26.24 & 64.91 & 27.99 \\
    0.40 & 13.10 & 38.70 & 41.40 & 37.34 & 88.43 & 37.70 \\
    0.50 & 18.25 & 47.63 & 55.80 & 47.06 & 99.97 & 42.80 \\
    \bottomrule
  \end{tabular*}
\end{table}

\clearpage

\subsection{Supervised combination of uncertainty scores}
\label{appendix:supervised-baseline}
COSMIC~\cite{hoffmann2022high} derives its confidence score from a supervised model trained on features
of the annotation pipeline. To test the value of supervised feature combination in our setting, we fit
a separate logistic-regression model for each architecture and $K\in\{1,5,20\}$. The binary target is
whether the reference molecule occurs in the top-$K$ set. As features, the model uses: highest candidate score
$s_{(1)}$, the top-$1$ score gap, the cutoff-specific boundary gap
$s_{(K)}-s_{(K+1)}$, $\log |\mathcal{C}|$, confidence, normalized entropy, cutoff-matched rank variance,
retrieval-level total, aleatoric and epistemic uncertainty, the predicted- and top-candidate fingerprint
cardinalities and their absolute difference, precursor $m/z$, and peak count. Softmax-derived features
use $T_{\mathrm{eval}}=0.003$.
All preprocessing and model selection are confined to the official validation fold. The
inverse regularization strength is selected by mean validation log loss. The model is then refit on the complete validation fold.
\begin{table}[htbp]
  \color{blue}
  \centering
  \caption{\rev{Comparison of relAURC values for the supervised logistic score described in Section \ref{appendix:supervised-baseline} with the strongest individual score.
  A separate model is trained for each architecture and retrieval cutoff. Lower relAURC is better.
  $\Delta$ is the relAURC of the best individual score minus that of the score combination, so positive
  values favor supervised combination. }}
  \label{tab:supervised-combination}
  \footnotesize
  \setlength{\tabcolsep}{4pt}
  \renewcommand{\arraystretch}{1.05}
  \begin{tabular*}{\textwidth}{@{\extracolsep{\fill}}lll rrr@{}}
    \toprule
    Architecture & $K$ & Best individual score & Individual relAURC & relAURC of combination & $\Delta$ \\
    \midrule
    MLP & 1  & $\kappa_{\mathrm{ret}}^{\mathrm{tot}}$ & 0.347 & 0.362 & $-0.014$ \\
    MLP & 5  & $\kappa_{\mathrm{ret}}^{\mathrm{tot}}$ & 0.323 & 0.306 & 0.017 \\
    MLP & 20 & $\kappa_{\mathrm{rank}}$ ($K\!=\!5$) & 0.322 & 0.277 & 0.045 \\
    Transformer & 1  & $\kappa_{\mathrm{conf}}$ & 0.356 & 0.363 & $-0.007$ \\
    Transformer & 5  & $\kappa_{\mathrm{ret}}^{\mathrm{tot}}$ & 0.317 & 0.297 & 0.019 \\
    Transformer & 20 & $\kappa_{\mathrm{ret}}^{\mathrm{tot}}$ & 0.320 & 0.275 & 0.044 \\
    \bottomrule
  \end{tabular*}
\end{table}
In Table \ref{tab:supervised-combination}, we see that for this setting, a supervised combination does not
 improve hit rate for $K=1$: the relAURC values are slightly
higher than that of the strongest individual score for both architectures.
It however provides modest gains for Hit@$5$ and larger gains for Hit@$20$, where
relAURC decreases by approximately $0.045$ for both the MLP and transformer.
This consistent pattern suggests that the scores and context features contain
complementary information for larger top-$K$ decisions, whereas a single
score already captures most of the useful signal for top-$1$ confidence.
Supervised combination can therefore be valuable when representative labeled
validation data are available, but it is not uniformly better than a strong
individual criterion.

\subsection{Sensitivity to the softmax temperature}
\label{appendix:temperature}
\begin{figure}[htbp]
  \color{blue}
  \centering
  \includegraphics[width=\textwidth]{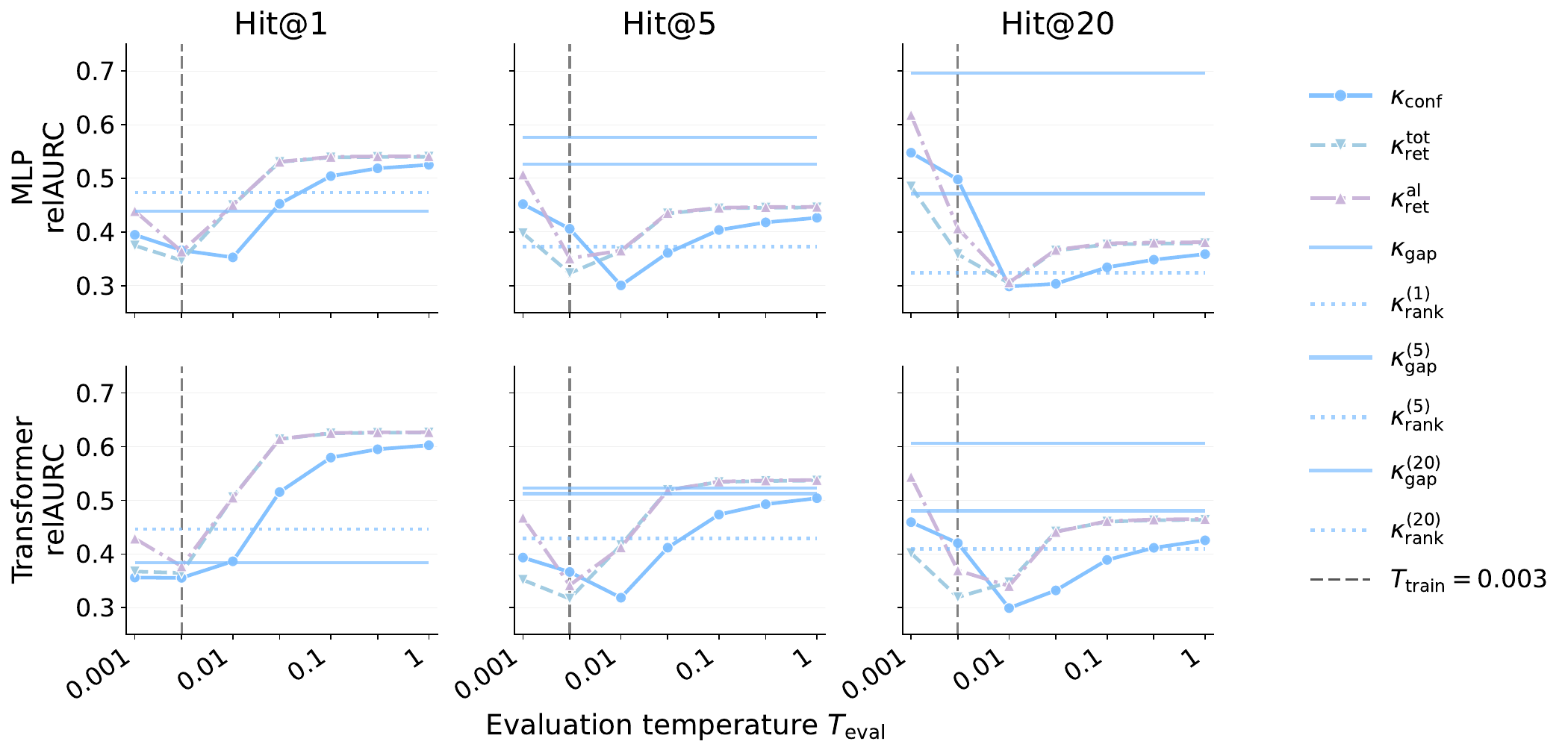}
  \caption{\rev{Sensitivity of selective prediction quality (relAURC, lower is better) to the evaluation
  temperature $T_{\mathrm{eval}}$ of the candidate softmax, for the MLP model (top) and the transformer model (bottom)
  Deep Ensembles and $\ell_K = 1 - \mathrm{Hit@}K$, $K\in\{1,5,20\}$, on the MassSpecGym test set.
  Dashed curves are the temperature-dependent uncertainty scores, solid lines are the temperature-independent scores. The dashed vertical
   line marks the training temperature $T_{\mathrm{train}}=0.003$.}}
  \label{fig:temp-sensitivity}
\end{figure}
Several retrieval-level scores are computed from the temperature-scaled candidate softmax (Eqn.~\ref{eq:candidate_prob}) and
therefore depend on the temperature $T_{\mathrm{eval}}$ at which it is evaluated.
Figure~\ref{fig:temp-sensitivity} shows their relAURC as a function of $T_{\mathrm{eval}}$ for both
architectures, together with the temperature-independent scores (score gap and rank variance).
 The softmax-based scores show a
clear optimum, whose location depends on the retrieval cutoff $K$. It lies close to
the training temperature $T_{\mathrm{train}}=0.003$ for Hit@1 and shifts to higher temperatures for
Hit@5 and Hit@20. These scores can thus be improved by tuning the evaluation temperature, whereas the score gap and rank variance are unaffected by $T$ and
require no such tuning.

\subsection{Alternative candidate sets}
\label{appendix:candidate-sets}
\begin{figure}[htbp]
\color{blue}
\centering
\includegraphics[width=\textwidth]{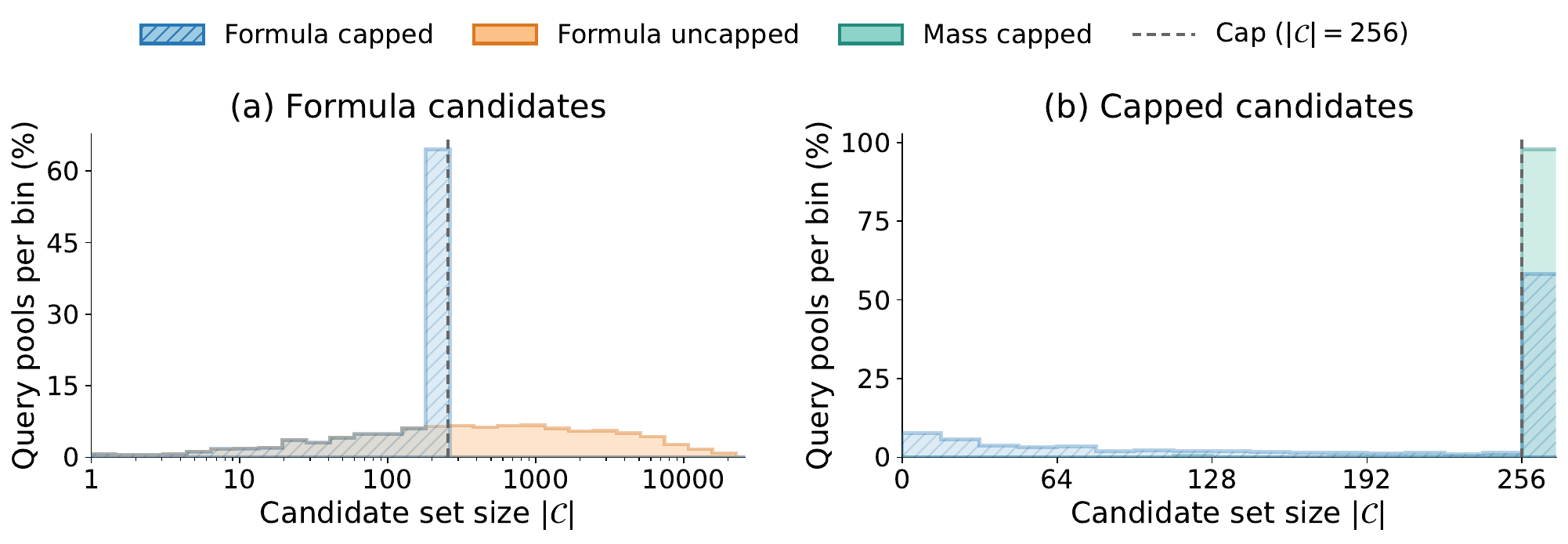}
\caption{\rev{Distribution of candidate-set sizes for the matched query pools used in the candidate-setting
evaluation. \textbf{Left}: formula-filtered candidate sets. Uncapped PubChem pools extend beyond the
MassSpecGym limit of 256 candidates; the horizontal axis is logarithmic. \textbf{Right}: comparison of
capped candidate sets based on molecular formula (blue) and precursor mass (turquoise).}}
\label{fig:candidate-sets}
\end{figure}
The MassSpecGym benchmark caps candidate sets at 256 entries. We examine two
alternative settings. First, the MLP and transformer ensembles trained on
the official capped formula setting are evaluated on matched capped and
uncapped formula-based PubChem pools. Second, an MLP ensemble is trained and
evaluated using candidates filtered by precursor mass. All comparisons use the same
spectra. Formula-based candidate sets contain compounds with the same elemental formula as the query.
 Mass-based candidate sets contain compounds whose neutral monoisotopic mass is within \(10\) ppm of the query mass
  inferred from the precursor \(m/z\), charge, and adduct.

Removing the cap on the number of candidates substantially increases retrieval difficulty.
In the uncapped dataset, approximately $59\%$ of the formula pools contain more than 256 candidates,
with some containing more than $20{,}000$. See Figure~\ref{fig:candidate-sets} for a comparison of distributions.
 Consequently, Hit@$K$ decreases
for both architectures.
Nevertheless, the ordering of the scoring functions remains stable across both architectures. Total
retrieval uncertainty remains the strongest criterion, confidence and rank variance remain strong alternatives. The principal
conclusions therefore extend to substantially larger candidate pools.
\begin{table}[htbp]
  \color{blue}
  \centering
  \caption{\rev{Retrieval accuracy under alternative candidate constructions.
  The formula models are trained on the official capped formula candidates; the paired capped and
  uncapped evaluations are constructed from the same PubChem candidate pools and use the same query
  mask. The mass-based results use an MLP trained and evaluated on capped mass-filtered candidates.
  Hit@$K$ values are percentages.}}
  \label{tab:candidate-setting-results}
  \footnotesize
  \setlength{\tabcolsep}{4pt}
  \renewcommand{\arraystretch}{1.05}
  \begin{tabular*}{\textwidth}{@{\extracolsep{\fill}}lll rrr@{}}
    \toprule
    Arch. & Train & Evaluation
      & H@1 & H@5 & H@20 \\
    \midrule
    MLP & Formula cap. & Formula paired cap.
      & 13.19 & 27.64 & 47.68 \\
    MLP & Formula cap. & Formula uncapped
      & 11.53 & 22.97 & 40.09 \\
    Transformer & Formula cap. & Formula paired cap.
      & 17.82 & 35.93 & 56.88 \\
    Transformer & Formula cap. & Formula uncapped
      & 14.74 & 28.77 & 47.65 \\
    MLP & Mass cap. & Mass cap.
      & 14.06 & 27.65 & 45.22 \\
    \bottomrule
  \end{tabular*}
\end{table}
Note that the lower relAURC values obtained after uncapping do not indicate improved
retrieval accuracy. Rather, they show that the scoring functions recover a
larger fraction of the achievable selective-prediction improvement. This is
consistent with the greater variation in candidate-set size providing a more
informative difficulty signal.
\begin{table}[htbp]
  \color{blue}
  \centering
  \caption{\rev{Selective-prediction quality under alternative candidate
  constructions, measured by relAURC (lower is better). Formula models are
  trained on the official capped formula candidates, whereas the mass-based
  MLP is trained on capped mass-filtered candidates. The paired capped and uncapped
  formula evaluations use matched query pools. Rank variance is matched to
  the evaluated cutoff $K$, and $-\log|\mathcal{C}|$ denotes the
  confidence-oriented candidate-set-size criterion. All softmax-derived
  scores use $T_{\mathrm{eval}}=0.003$. The best value in each row is shown
  in bold.}}
  \label{tab:candidate-setting-relaurc}
  \scriptsize
  \setlength{\tabcolsep}{2.5pt}
  \renewcommand{\arraystretch}{1.05}
  \begin{tabular*}{\textwidth}{@{\extracolsep{\fill}}llr rrrrrrr@{}}
    \toprule
    Arch. & Evaluation candidates & $K$
      & $\kappa_{\mathrm{conf}}$
      & $\kappa_{\mathrm{gap}}$
      & $\kappa_{\mathrm{ret}}^{\mathrm{tot}}$
      & $\kappa_{\mathrm{ret}}^{\mathrm{al}}$
      & $\kappa_{\mathrm{ret}}^{\mathrm{ep}}$
      & $\kappa_{\mathrm{rank}}^{(K)}$
      & $-\log|\mathcal{C}|$ \\
    \midrule
    \multirow{3}{*}{MLP} & \multirow{3}{*}{Formula paired cap.}
      & 1  & 0.363 & 0.437 & \textbf{0.346} & 0.362 & 0.431 & 0.494 & 0.550 \\
    & & 5  & 0.394 & 0.579 & \textbf{0.315} & 0.346 & 0.442 & 0.363 & 0.461 \\
    & & 20 & 0.485 & 0.694 & 0.355 & 0.403 & 0.483 & \textbf{0.328} & 0.391 \\
    \addlinespace
    \multirow{3}{*}{MLP} & \multirow{3}{*}{Formula uncapped}
      & 1  & 0.318 & 0.409 & \textbf{0.303} & 0.318 & 0.395 & 0.402 & 0.469 \\
    & & 5  & 0.362 & 0.559 & \textbf{0.282} & 0.301 & 0.425 & 0.319 & 0.334 \\
    & & 20 & 0.427 & 0.684 & 0.269 & 0.305 & 0.405 & \textbf{0.228} & 0.254 \\
    \addlinespace
    \multirow{3}{*}{Transformer} & \multirow{3}{*}{Formula paired cap.}
      & 1  & \textbf{0.346} & 0.376 & 0.361 & 0.371 & 0.503 & 0.429 & 0.645 \\
    & & 5  & 0.369 & 0.520 & \textbf{0.320} & 0.348 & 0.536 & 0.434 & 0.552 \\
    & & 20 & 0.421 & 0.610 & \textbf{0.321} & 0.367 & 0.591 & 0.409 & 0.480 \\
    \addlinespace
    \multirow{3}{*}{Transformer} & \multirow{3}{*}{Formula uncapped}
      & 1  & \textbf{0.314} & 0.365 & 0.325 & 0.336 & 0.451 & 0.390 & 0.536 \\
    & & 5  & 0.333 & 0.507 & \textbf{0.287} & 0.303 & 0.489 & 0.375 & 0.395 \\
    & & 20 & 0.366 & 0.626 & \textbf{0.255} & 0.290 & 0.483 & 0.288 & 0.321 \\
    \addlinespace
    \multirow{3}{*}{MLP} & \multirow{3}{*}{Mass capped}
      & 1  & 0.497 & 0.572 & \textbf{0.479} & 0.488 & 0.576 & 0.620 & 0.969 \\
    & & 5  & 0.523 & 0.670 & \textbf{0.459} & 0.475 & 0.590 & 0.670 & 0.992 \\
    & & 20 & 0.608 & 0.745 & \textbf{0.531} & 0.554 & 0.651 & 0.742 & 1.021 \\
    \bottomrule
  \end{tabular*}
\end{table}
The mass-based setting yields similar Hit@$1$ and Hit@$5$ but lower Hit@$20$ than
the capped formula setting. Total retrieval uncertainty is the
strongest criterion at every cutoff, although selective ordering is generally
weaker. Nearly all mass-filtered candidate sets reach the cap, making
candidate count uninformative and reducing the relative advantage of rank
variance. Thus, total retrieval uncertainty is robust across candidate
definitions, whereas the usefulness of candidate-size and rank-stability
criteria depends more strongly on how the candidate pool is constructed.

\end{revblock}

\end{appendices}

\clearpage
\twocolumn
\bibliography{sn-bibliography}

\end{document}